\documentclass[journal,compsoc]{IEEEtran}
\pdfoutput=1

\usepackage{amsmath}
\usepackage{multirow}
\usepackage{siunitx}
\usepackage{graphicx}
\usepackage{stfloats}
\usepackage{booktabs}
\usepackage{caption}
\usepackage{subcaption}
\usepackage{textcomp}
\usepackage[hidelinks]{hyperref}
\usepackage{float}
\usepackage[backend=biber,bibstyle=ieee,citestyle=numeric-comp,maxbibnames=2,minbibnames=1,mincitenames=1,maxcitenames=2,doi=false,isbn=false, url=false]{biblatex}
\usepackage{xcolor}
\usepackage{xspace}

\renewcommand{\appendixname}{Supplementary}

\newcommand{\DAQAp}{DAQA$^\prime$\xspace}
\newcommand{\VisualFiLM}{\textbf{\texttt{Visual FiLM Resnet101}}\xspace}
\newcommand{\NAAQATwoDConv}{\textbf{\texttt{NAAQA 2D Conv}}\xspace}
\newcommand{\NAAQAParallel}{\textbf{\texttt{NAAQA Parallel}}\xspace}
\newcommand{\NAAQAInterleavedTime}{\textbf{\texttt{NAAQA Interleaved Time}}\xspace}
\newcommand{\NAAQAInterleavedFreq}{\textbf{\texttt{NAAQA Interleaved Freq}}\xspace}
\newcommand{\NAAQAParallelOptimized}{\textbf{\texttt{Optimized NAAQA Parallel}}\xspace}
\newcommand{\NAAQATwoDConvOptimized}{\textbf{\texttt{Optimized NAAQA 2D Conv}}\xspace}
\newcommand{\MALiMo}{\textbf{\texttt{MALiMo ctrl}}\xspace}

\begin{document}
\title{NAAQA: A Neural Architecture for\\Acoustic Question Answering}

\author{
Jérôme~Abdelnour, Jean~Rouat~\IEEEmembership{Senior Member,~IEEE}, and Giampiero~Salvi~\IEEEmembership{Member,~IEEE}
\thanks{J. Abdelnour and J. Rouat are with NECOTIS, Dept. of Electrical and Computer Engineering, Sherbrooke University \{Jerome.Abdelnour,Jean.Rouat\}@usherbrooke.ca.}
\thanks{G. Salvi is with Department of Electronic Systems, Norwegian University of Science and Technology giampiero.salvi@ntnu.no, and with KTH Royal Institute of Technology, Dept. of Electrical Engineering and Computer Science.}
}

\markboth{IEEE Transactions on Pattern Analysis and Machine Intelligence, VOL 45, ISSUE 4, Pages 4997--5009, DOI: \href{https://doi.org/10.1109/TPAMI.2022.3194311}{10.1109/TPAMI.2022.3194311}}{}

\IEEEtitleabstractindextext{
    \begin{abstract}
    The goal of the Acoustic Question Answering (AQA) task is to answer a free-form text question about the content of an acoustic scene. It was inspired by the Visual Question Answering (VQA) task. 
    In this paper, based on the previously introduced CLEAR dataset, we propose a new benchmark for AQA, namely CLEAR2, that emphasizes the specific challenges of acoustic inputs.
    These include handling of variable duration scenes, and scenes built with elementary sounds that differ between training and test set.
    We also introduce NAAQA, a neural architecture that leverages specific properties of acoustic inputs. 
    The use of 1D convolutions in time and frequency to process 2D spectro-temporal representations of acoustic content shows promising results and enables reductions in model complexity. 
    We show that time coordinate maps augment temporal localization capabilities which enhance performance of the network by $\sim$17 percentage points. On the other hand, frequency coordinate maps have little influence on this task.
    NAAQA achieves 79.5\% of accuracy on the AQA task with $\sim$4 times fewer parameters than the previously explored VQA model.
    We evaluate the performance of NAAQA on an independent data set reconstructed from DAQA.
    We also test the addition of a MALiMo module in our model on both CLEAR2 and DAQA.
    We provide a detailed analysis of the results for the different question types.
    We release the code to produce CLEAR2 as well as NAAQA to foster research in this newly emerging machine learning task.
\end{abstract}

}

\maketitle
\IEEEdisplaynontitleabstractindextext

\IEEEpeerreviewmaketitle

\section{Introduction}
\IEEEPARstart{Q}{uestion} answering (QA) tasks are examples of constrained and limited scenarios for research in reasoning. The agent's task in QA is to answer questions based on context. Text-based QA uses text corpora as context~\cite{voorhees1999trec, voorhees2000building, soubbotin2001patterns, hovy2000question, iyyer2014neural,ravichandran2002learning}. In visual question answering (VQA) the questions are related to a scene depicted in still images~\cite{johnson2017clevr,antol2015vqa,zhu2016visual7w,gao2015you, agrawal2016analyzing, zhang2016yin, geman2015visual}. Finally, video question answering attempts to use both the visual and acoustic information in video material as context~\cite{cao2005automated,chua2003question,yang2003videoqa,kim2017deepstory,tapaswi2016movieqa,wu2008robust}. The use of the acoustic channel is usually limited to linguistic information that is expressed in text form, either with manual transcriptions (e.g. subtitles) or by automatic speech recognition~\cite{ZhangEtAl2017VQAwithspeech}.

In most studies, reasoning is supported by spatial and symbolic representations in the visual domain~\cite{chang1996symbolic,moktefi2013visual}. However, reasoning and logic relationships can also be studied via representations of sounds~\cite{champagne2015}. Including the auditory modality in studies on reasoning is of particular interest for research in artificial intelligence \cite{audioretrieval2022}, but also has implications in real world applications~\cite{champagne2018}.
In \cite{PieropanEtAl2014IROS}, audio was used in combination with video and depth information to recognize human activities. It was shown that sound can be more discriminative than the corresponding visual cues. As an example, imagine using an espresso machine. Besides possibly a display, all information about the different phases of producing coffee, from grinding the beans, to pressing the powder into the holder and brewing the coffee with high pressure hot water are conveyed by the sounds. 
Detection of abnormalities in machinery where the moving parts are hidden, or the detection of threatening or hazardous events are other examples of the importance of the audio information for cognitive systems.

The audio modality provides important information that can be leveraged in the context of QA reasoning. Audio allows QA systems to answer relevant questions more accurately, or even to answer questions that are not approachable from the visual domain alone.
In \cite{CleararXiv2018}, we introduced the AQA task and proposed a new database (CLEAR) to promote research in AQA. The agent's goal, in the proposed task, was to answer questions related to \emph{acoustic scenes} composed by a sequence of \emph{elementary musical sounds}. The questions foster reasoning on the properties of the elementary sounds and their relative and absolute position in the scene. 
To build CLEAR, we were inspired by the work of Johnson \emph{et al.} \cite{johnson2017clevr} for VQA. Similarly, we tested an architecture built for VQA and based on \emph{FiLM layers}~\cite{perez2017film} on the newly proposed AQA task.
\citeauthor{Fayek2019} \cite{Fayek2019} later proposed to extend the questions to more acoustically realistic situations by developing a new database called DAQA. To evaluate the results, they proposed the MALiMo network which relies on several FiLM layers.

The works cited above use neural network architectures that are largely inspired by image processing research.
However, the structure of acoustic data is fundamentally different from that of visual data. This is illustrated for example in~\cite{ZhangEtAl2019AAAI} where two standard data sets in computer vision (MNIST) and speech technology (Google Speech Commands) are compared via T-SNE~\cite{Maaten2008}.
A legitimate question is whether it is possible to obtain better results (in terms of accuracy and network complexity) by adapting the first layers of the architectures to take into account intrinsic characteristics of acoustic signals.
Even within the AQA domain, the properties of acoustic data may vary significantly depending on the nature of the auditory scenes (e.g. CLEAR vs DAQA). It is, therefore interesting to evaluate the impact of the dataset on system performance.

To answer the above questions, we present a study that evaluates the impact of audio pre-processing, of acoustic feature extraction and of dataset characteristics on the performance neural architectures for AQA.
When considering performance, we focus both on accuracy and complexity of the models.
We provide a detailed analysis of our results based on question type to improve interpretability.
The main contributions can be summarized as follows:


\vspace{-0.2em}
\begin{itemize}
    \item We introduce CLEAR2 a more challenging version of the CLEAR dataset, which comprises scenes of variable duration and different elementary sounds for the training and test sets.  
    \item We propose a highly optimized FiLM-based architecture (NAAQA) inspired by VQA tasks containing new feature extraction modules that are tailored to acoustic inputs.
    \item We study the effect of time and frequency coordinate maps for acoustic data at different levels in the architecture.
    \item We evaluate the generality of the methods by testing NAAQA on a regenerated a version of the DAQA dataset (\DAQAp) and by adding a MALiMo module (from \cite{Fayek2019}) into our NAAQA architecture.
    \item We provide a detailed analysis of our experimental results that helps interpretability of the model.
\end{itemize}
On the CLEAR2 dataset NAAQA outperforms the VQA baseline (which is 4 times more complex in terms of number of parameters) by 17.2 percent points in the accuracy score.

The rest of the paper is organized as follows: Section~\ref{sec:related_work} reports on recent related work, Section~\ref{datasets} describes both our CLEAR2 dataset and the \DAQAp dataset, Section~\ref{sec:method} presents the QA models we have tested, Section~\ref{sec:experiments} gives details on the experimental settings, Section~\ref{sec:results_and_discussion} presents and discusses the results and, finally, Section~\ref{sec:conclusions} concludes the paper. Some extra information is reported in the supplementary material. 

\section{Related Work}
\label{sec:related_work}
This section presents previous research in QA systems including data generation and modeling.
\subsection{Text-Based Question Answering}
The question answering task was introduced as part of the Text Retrieval Conference \cite{voorhees1999trec}. 
In text-based question answering, both the questions and the context are expressed in text form.
Answering these questions can often be approached as a pattern matching problem in the sense that the information can be retrieved almost verbatim in the text (e.g. \cite{soubbotin2001patterns, hovy2000question, ravichandran2002learning, iyyer2014neural}). 

\subsection{Visual Question Answering (VQA)}
Visual Question Answering aims to answer questions based on a visual scene.
Several VQA datasets are available to the scientific community~\cite{johnson2017clevr, Hudson_2019_CVPR,wang_FVQA,Marino_2019_CVPR,zellers2019vcr, Gordon_2018_CVPR,zhu2016visual7w,gao2015you,balanced_vqa_v2,antol2015vqa}.
However, designing an unbiased dataset is non-trivial.
\citeauthor{agrawal2016analyzing} \cite{agrawal2016analyzing} observed that the type of questions has a strong impact on the results of neural network based systems which motivated research to reduce the bias in VQA datasets~\cite{Manjunatha_2019_CVPR,das_bias,Agrawal_2018_CVPR,zhang2016yin, geman2015visual, johnson2017clevr}. 
Gathering good labeled data is also non-trivial which induced \citeauthor{zhang2016yin} and \citeauthor{geman2015visual} \cite{zhang2016yin, geman2015visual} to constrain their work to yes/no questions. 
To alleviate this problem, \citeauthor{johnson2017clevr} \cite{johnson2017clevr} proposed the use of synthetic data for both questions and visual scenes. 
The resulting CLEVR dataset has been extensively used to evaluate neural networks for VQA applications \cite{perez2017film, hudson2018compositional, prob_clevr2019, Hu_2019_ICCV, hu_stack_clevr_2018, NIPS2018_YI} which helped foster research on VQA.
To create visual scenes, the authors automated a 3D modelling software.
This allows for an unlimited supply of labeled data eliminating the time and effort needed for manual annotations.
For the questions, they first manually designed semantic representations for each type of question. 
These representations describe the reasoning steps needed to answer a question (i.e. ``find all cubes $|$ that are red $|$ and metallic''). 
The semantic representations are then instantiated based on the visual scene composition thus creating a question and an answer for a given scene.
This gives complete control over the labelling process. 



\subsection{Dababases for AQA}
\begin{figure*}
  \includegraphics[width=\linewidth]{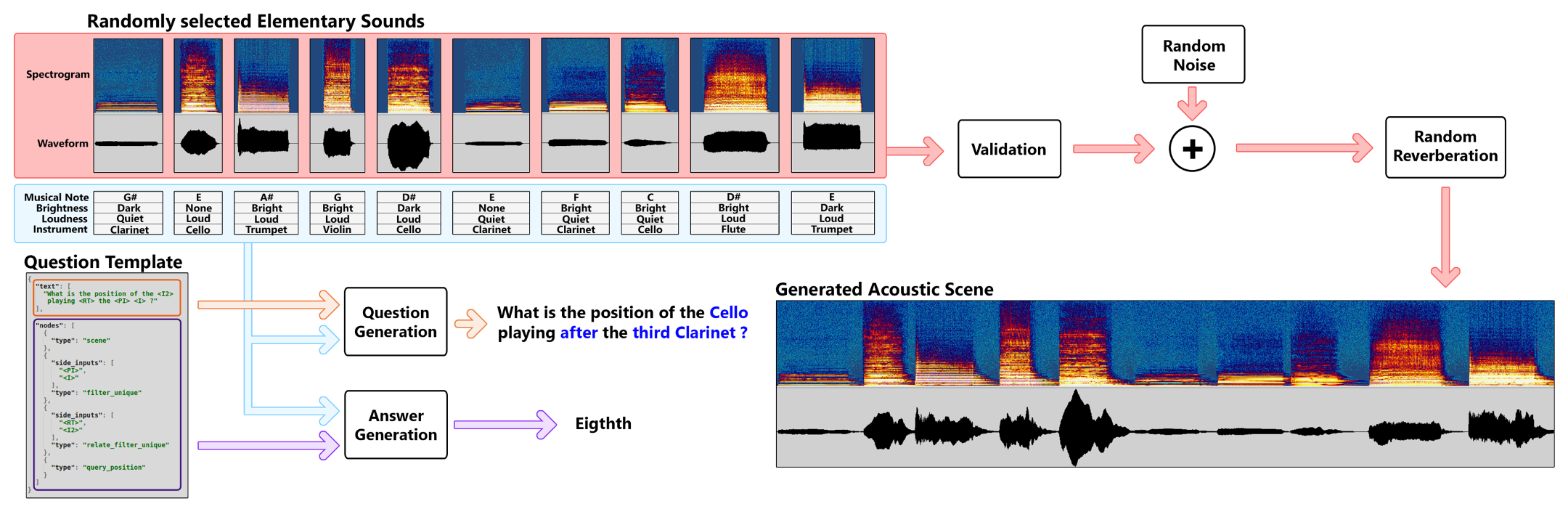}
  \caption{\textbf{Overview of the CLEAR dataset generation process.} Highlighted in red: 10 randomly sampled sounds from the elementary sounds bank, are assembled to create an acoustic scene. The attributes of each elementary sound are depicted in blue. The question template (orange) and the elementary sounds attributes are combined to instantiate a question.
  The answer is generated by applying each steps of the question functional program (purple) on the acoustic scene definition (blue).
  The impact of the reverberations can be seen in the changes of the signals envelops.}
  \label{fig:generation_process}
  \vspace{-1em}
\end{figure*}
As in VQA, using generated data in the design of AQA datasets has substantial advantages.
Data can be automatically annotated which saves time and complexity.
The number of training examples that can be generated is only limited by the available computational resources.
Controlling the generation process gives a complete understanding of the properties and relations of the objects in a scene.
This understanding can be leveraged to reduce bias in the dataset and to generate complex questions and their corresponding answers.
The CLEAR dataset ~\cite{CleararXiv2018} has been initially generated using semi-synthetic data.
The elementary sounds were real recordings of musical notes played by various instruments and players.
The auditory scenes were obtained by concatenating these elementary sounds in different combinations.
The data set had two main limitations: scenes had fixed duration, and the same elementary sounds were used to generate the test and training scenes (although test and training scenes were different).
The DAQA dataset~\cite{Fayek2019} comprises more complex and less stationary elementary natural sounds coming for example from aircrafts, cars, doors, human speaking, bird singing, dog barking, etc. Although more complex and varied than CLEAR, the evaluation also uses the same elementary sound recordings for training and testing.

In this paper, we propose a more challenging version of CLEAR which uses different elementary sound recordings for the training and test sets and generates variable duration auditory scenes. 

\subsection{Convolutional neural network on Audio}
Convolutional neural networks (CNN) have dominated the visual domain in recent years.
More recently, they have also been applied to a number of problems in the acoustic domains such as acoustic scene classification \cite{abdoli2019endtoend, Boddapati2017, lee2017raw, Han2016AcousticScene}, music genre classification \cite{Nam2019MusicGenre, lee2017raw}, instrument classification \cite{zheng2018cnnsbased, pons2017timbre}, sound event classification and localization \cite{brousmiche2020soundClassification} and speech recognition \cite{lee2017raw}.
Some authors~\cite{Nam2019MusicGenre, zheng2018cnnsbased, pons2017timbre, Han2016AcousticScene} use intermediate representations such as STFT~\cite{stft1987}, MFCC~\cite{mfcc2000} or CQT~\cite{CQT1992} spectrograms while others work directly with the raw audio signal~\cite{lee2017raw, abdoli2019endtoend}.

Square convolutional and pooling kernels are often used to solve visual task such as VQA, visual scene classification and object recognition \cite{krizhevsky2012imagenet, simonyan2014deep, he2016deep, szegedy2014going}. 
\citeauthor{Boddapati2017, cnnAudioClass2017, kumar2017deep} \cite{Boddapati2017, cnnAudioClass2017, kumar2017deep} have successfully used visually motivated CNN with square filters to solve audio related tasks. 
Time-frequency representations of audio signals are however structured very differently than visual representations. 
\citeauthor{Pons2016} \cite{Pons2016} explore the performance of different structures of convolutive kernels when working with music signals classification. They propose the use of 1D convolution kernels to capture time-specific or frequency-specific features. They demonstrate that similar accuracy can be reached using a combination of 1D convolutions instead of 2D convolutions by combining 1D time and 1D frequency filters while using much fewer parameters. They also explore rectangular kernels which capture both time and frequency features at different scales. The impact of such strategies for music classification 
is still an open question in the context of auditory scene analysis.

Coordinate maps initially proposed in~\cite{liu2018intriguing} by~\citeauthor{liu2018intriguing} have proven successful for processing visual data in the context of VQA. The method consists in augmenting the visual input with matrices containing numbers in the range -1 to 1 which vary either in the $x$ or in the $y$-dimension. With MALiMo~\cite{Fayek2019}, the same strategy is used to indicate the simultaneous relative positions of features in frequency and time.
\citeauthor{farnet2019} \cite{farnet2019} proposed \emph{Frequency-Aware convolutions} which are equivalent to concatenating coordinate maps only in the \emph{frequency} axis.
The effectiveness of coordinate maps on the time dimension for audio signals have not been studied to the best of our knowledge. 



In this study we first evaluate the performance of a network initially designed for the VQA task (Visual FiLM) \cite{perez2017film} on the AQA task, using the CLEAR2 data set.
Then we introduce the NAAQA architecture to leverage specific properties of acoustic inputs.
For this architecture, we analyze the influence of separate time and frequency coordinate maps.
We then study the impact of adding a MALiMo block into our architecture.
Finally, we evaluate our model on the \DAQAp dataset.


\vspace{-3mm}
\section{Data}
\label{datasets}
We use two very different datasets in our experiments in order to study the effect of the AQA task characteristics on the model performance.
The first set, that is also a contribution of this paper, comprises musical sounds (CLEAR2); the second includes short environmental sounds (\DAQAp). 
\subsection{CLEAR2}
\label{sec:dataset}
CLEAR2 is an updated version of CLEAR~\cite{CleararXiv2018}. 
A graphical overview of the generation process is depicted in Figure~\ref{fig:generation_process}. 
Each record in the dataset is a unique combination of a scene, a question and an answer.
\newcommand{\emphPH}[1]{\textit{\textbf{#1}}}
\begin{table*}
    \centering
    \resizebox{\textwidth}{!}{
    \footnotesize
    \begin{tabular}{lp{0.5\textwidth}p{0.3\textwidth}c}
    \hline\hline
    Question type & Example & Possible Answers & \# \\
    \hline
    Note         & What is the note played by the \emphPH{flute} that is \emphPH{after} the \emphPH{loud} \emphPH{bright} \emphPH{D} note? & A, A\#, B, C, C\#, D, D\#, E, F, F\#, G, G\# & 12\\
    Instrument   & What instrument plays a \emphPH{dark} \emphPH{quiet} sound in the \emphPH{end} of the scene? & bass, cello, clarinet, flute, trumpet, violin & 5 \\
    Brightness & What is the brightness of the \emphPH{first} \emphPH{clarinet} sound? & bright, dark & 2 \\
    Loudness & What is the loudness of the \emphPH{violin} playing \emphPH{after} the \emphPH{third} \emphPH{trumpet}? & quiet, loud & 2 \\
    Absolute Position & What is the position of the \emphPH{A\#} note playing \emphPH{before} the \emphPH{bright} \emphPH{B} note? & \multirow{2}{*}{$\Big\}$ first, second ... fifteenth} & \multirow{2}{*}{15} \\
    Relative Position & Among the \emphPH{trumpet} sounds which one is a \emphPH{F}? &  &  \\
    Global Position & In what part of the scene is the \emphPH{clarinet} playing a \emphPH{loud} \emphPH{G} note ? & beginning, middle, end (of the scene) & 3 \\
    Counting     & How many other sounds have the same brightness as the \emphPH{third} \emphPH{violin}? & \multirow{2}{*}{$\Big\}$0, 1 ... 15} & \multirow{2}{*}{16} \\
    Counting Instruments & How many different instruments are playing \emphPH{before} the \emphPH{second} \emphPH{trumpet}? &  & \\
    Exist        & Is there a \emphPH{bass} playing a \emphPH{bright} \emphPH{C\#} note? & \multirow{2}{*}{$\Big\}$yes, no} & \multirow{2}{*}{2}\\
    Counting comparison & Is there an \emphPH{equal} number of \emphPH{loud} \emphPH{cello} sounds and \emphPH{quiet} \emphPH{clarinet} sounds? &  & \\
    
    \hline
    Total & & & 57 \\
    \hline\hline
    \end{tabular}
    }
    \caption{\textbf{Types of questions with examples and possible answers.} The variable parts of each question is emphasized in bold italics. 
    The number of possible answer per question type is reported in the last column. Certain questions have the same possible answers, the meaning of which depends on the type of question.}
    \label{tab:types_of_question}
    \vspace{-1em}
\end{table*}
To build acoustic scenes, we prepared a bank of elementary sounds composed of real musical instrument recordings extracted from the Good-Sounds~\cite{Bandiera2016} dataset\footnote{
Each elementary sound in a scene is characterized by an n-tuple: [\emph{Instrument}, \emph{Brightness}, \emph{Loudness}, \emph{Musical Note}, \emph{Duration}, \emph{Absolute position in scene}, \emph{Relative position in scene}, \emph{Global position}].
The \emph{Brightness} property is computed by using the \texttt{timbralmodels} \cite{timbral_model} library. 
A threshold is used to define the label of the sound (\emph{Dark} or \emph{Bright}).
The \emph{Loudness} labels are assigned based on the perceptual loudness as defined by the ITU-R BS.1770-4 international normalization standard \cite{ITULoudness2015}. 
Again, a threshold is used to determine if the sound is \emph{Quiet} or \emph{Loud}.
All attribute values are listed in Table \ref{tab:types_of_question} as possible answers to the questions explained below.}.
Differently from CLEAR, in CLEAR2 we make sure that the recordings (players, instruments, microphones) of the elementary sounds  are different for the training and test sets.
For the training set, the bank comprises 135 unique recordings (compared to 56 in CLEAR) sampled at 48KHz including 6 different instruments (bass, cello, clarinet, flute, trumpet and violin), 12 notes (chromatic scale) and 3 octaves. A different set of 135 recordings of the same instruments recorded using different microphones and players is used to create the test set.
The acoustic scenes are built by concatenating between 5 to 15 randomly chosen sounds from the elementary sound bank into a sequence (as opposed to CLEAR which comprised fixed duration scenes). Silence segments of random duration are added in-between elementary sounds. The acoustic scenes are then corrupted by filtering to simulate room reverberation and by adding a white uncorrelated uniform noise. Both the amount of noise and reverberation vary from scene to scene with the goal of increasing variability in the data.

For each scene, a number of questions is generated using CLEVR-like~\cite{johnson2017clevr} templates.
A template defines the reasoning steps required to answer a question based on the composition of the scene (i.e. ``find all instances of violin $|$ that plays before trumpet $|$ that is the loudest''). 
942 templates where designed for this AQA task.
Not all template instantiations results in a valid question. The generated questions are filtered to remove ill posed questions similarly to \cite{johnson2017clevr}. Table \ref{tab:types_of_question} shows examples of questions with their answers.

The a priori probability of answering correctly with no information about the question or the scene, and assuming a uniform distribution of classes, is $\frac{1}{57} = 1.75\%$.
These probabilities are higher, on average, if we introduce information about the question. For example, if we know that the question is of the type \emph{Exist} or \emph{Counting comparison}, there are only two possible answers (yes or no) and the probability of answering correctly by chance is $0.5$.
The majority class accuracy (always answering the most common answer: Yes) is 7.5\%.
Statistics on the CLEAR2 dataset are presented in Table~\ref{tab:dataset_statistics}.

The generation process was built with extensibility in mind.
Different versions of the dataset with fewer or more objects per scene can be generated by using different parameters for the generation script. It is also possible to modify the elementary sounds bank to generate datasets for AQA in other domains, speech or environmental sounds, for example. The code for generating the dataset is available on GitHub\footnote{\url{https://github.com/NECOTIS/CLEAR-AQA-Dataset-Generator}}. Pre-generated version of the dataset is available both on IEEE Dataport\footnote{\url{https://dx.doi.org/10.21227/7x26-a025}} and HuggingFace\footnote{\url{https://huggingface.co/datasets/J3romee/CLEAR}}.

\begin{table}
    \centering
    Datasets global statistics\\[1mm]
    \begin{tabular}{lcc}
        \hline
        \hline
                        & \multicolumn{2}{c}{Dataset} \\
                        & CLEAR2    & {\DAQAp} \\
        \hline
        \# of questions & 200 000   & 599 441 \\
        \# of scenes    & 50 000    & 100 000 \\
        \# of answers   & 57        & 52        \\
        \# of elementary sounds & 135 ( + 135 for test) & 358 \\
        \# of types of question & 11   &   5   \\
        \# of unique vocabulary words & 91  & 158 \\
    \hline
    \hline
    \end{tabular}
    
    \vspace{3mm}
    CLEAR2 Dataset detailed statistics\\[1mm]
    \begin{tabular}{lccc}
        \hline
        \hline
         & Mean & Min & Max \\
        \hline
        \# of sounds per scene & 10 & 5 & 15 \\
        Elementary sound duration &  0.85s & 0.69s & 1.11s \\
        Scene duration &  10.69s & 4.46s & 17.82s \\
        \# of words per question &  17 & 6 & 28 \\
        \# of unique words per question & 12 & 5 & 19 \\
    \hline
    \hline
    \end{tabular}
    
    \vspace{3mm}
    DAQA' Dataset detailed statistics\\[1mm]
    \begin{tabular}{lccc}
        \hline
        \hline
         & Mean & Min & Max \\
        \hline
        \# of sounds per scene & 9 & 5 & 12 \\
        Elementary sound duration &  9.35s & 0.6s & 20s \\
        Scene duration &  1min 19s & 9s & 3min 4s \\
        \# of words per question &  13 & 5 & 27 \\
        \# of unique words per question & 11 & 5 & 22 \\
    \hline
    \hline
    \end{tabular}
    \caption{\textbf{Datasets statistics}}
    \label{tab:dataset_statistics}
\end{table}

\subsection{Reproducing DAQA}
\label{data:reproducing_daqa}
We were not able to fully recreate the DAQA dataset because it relies on some AudioSet~\cite{audioset} YouTube videos that have since been deleted. We were able to retrieve 358 sounds out of the 400 sounds that were used to generate the original dataset.
We used these sounds to generate the dataset.
Changing the number of elementary sounds also impacts the whole generation process.
This dataset is therefore different from the original DAQA and will be referred to as {\DAQAp} from now on.
Our results are therefore not fully comparable to the ones reported in~\cite{Fayek2019}.
A list of all the missing sounds is available in Supplementary Materials.


\subsection{Comparing CLEAR2 and \DAQAp}
Table \ref{tab:dataset_statistics} report statistics on both CLEAR2 and \DAQAp.
The major difference between both datasets is the type of elementary sounds used to generate the acoustic scenes, that is, sustained musical notes (CLEAR2) versus possibily transient environmental sounds (\DAQAp).

Acoustic scenes in \DAQAp are much longer on average than the ones in CLEAR2. 
This results in much bigger input spectrograms, and, in turns, much higher computational requirements and longer training time.

Finally, the original DAQA~\cite{Fayek2019} and, consequently, our reconstruction (\DAQAp) suffer from the same problem as the original CLEAR.
The same elementary sounds are used in the training and test scenes.
Although scenes are still different between training and test set, this may cause the models to ``remember'' the elementary sounds rather than extracting their properties.
In CLEAR2, this problem was mitigated by using different elementary sounds for the training and test set.

\section{Method}
\label{sec:method}
We first describe the original Visual FiLM architecture \cite{perez2017film} that we use as baseline model, then the proposed modifications that lead to our NAAQA architecture and, finally, 
NAAQA with a MALiMo module.

\subsection{Baseline model: Visual FiLM}
\label{method:film_model}
\begin{figure}
  \begin{center}
      \includegraphics[width=0.95\columnwidth, keepaspectratio]{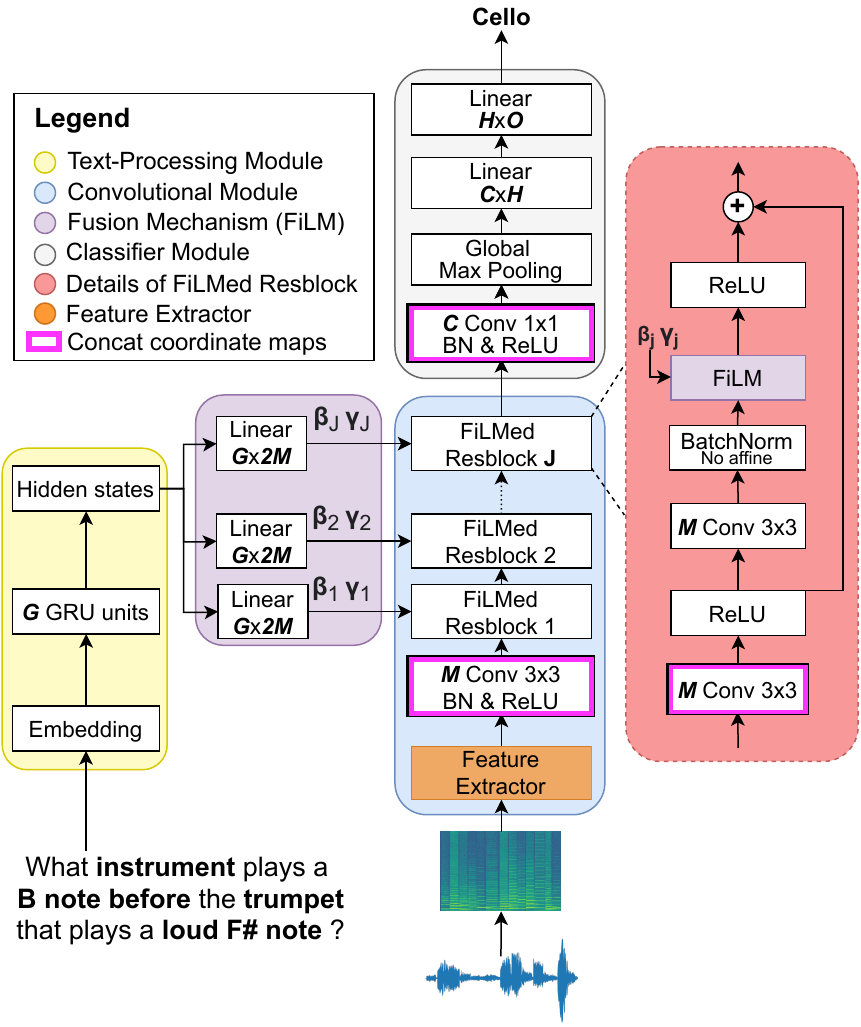}
      \caption{\textbf{Common Architecture.} Two inputs: a spectro-temporal representation of an acoustic scene and a textual question. 
      The spectro-temporal representation goes through a feature extractor (\emph{Parallel} and \emph{Interleaved} feature extractor detailed in Section \ref{sec_method_feature_extraction} for NAAQA and Resnet101 pretrained on ImageNet for Visual FiLM) and then a serie of $J$ Resblocks that are linearly modulated by $\boldsymbol\beta_j$ and $\boldsymbol\gamma_j$ (learned from the question input) via FiLM layers. Coordinate maps are inserted before convolution blocks that are illustrated with a pink border. The output is a probability distribution of the possible answers.}
      \label{fig:model_architecture}
  \end{center}
\end{figure}

Both the proposed NAAQA and Visual FiLM~\cite{perez2017film} share an overall common architecture which is depicted in Figure~\ref{fig:model_architecture}.
Visual FiLM, that we will use as baseline model, is inspired by Conditional Batch Normalization architectures~\cite{IoffeAndSzegedy2015ICMLConditionalBatchNormalization} and achieved state of the art results on the CLEVR VQA task~\cite{johnson2017clevr}. 
The network takes a visual scene and a text-based question as inputs and predicts an answer to the question for the given scene.
The text-processing module uses $G$ unidirectional gated recurrent units (GRUs) to extract context from the text input (yellow area in Figure~\ref{fig:model_architecture}).
The visual scene is processed by the convolutional module (blue area in the figure).
The first step of this module is feature extraction (orange box), performed by a Resnet101 model~\cite{he2016deep} pre-trained on ImageNet~\cite{russakovsky2015imagenet}.
The extracted features are processed by a convolutional layer with batch normalization \cite{IoffeAndSzegedy2015ICMLConditionalBatchNormalization} and ReLU \cite{relu} activation followed by $J$ Resblocks illustrated in details in the red area in the figure. Unless otherwise specified, batch normalization and ReLU activation functions are applied to all convolutional layers. Each Resblock $j$ comprises convolutional layers with $M$ filters that are linearly modulated by \emph{FiLM layers} through the two $M \times 1$ vectors $\boldsymbol\beta_j$ (additive) and $\boldsymbol\gamma_j$ (multiplicative). 
This modulation emphasizes the most important feature maps and inhibits the irrelevant maps given the context of the question. 
$\boldsymbol\beta_j$ and $\boldsymbol\gamma_j$ are learned via fully connected layers using the text embeddings extracted by the text processing module as inputs (purple area in the figure). 
The affine transformation in the batch normalization before the FiLM layer is deactivated.
The FiLM layer applies its own affine transformation using the learned $\boldsymbol\beta_j$ and $\boldsymbol\gamma_j$ to modulate features.
Several Resblocks can be stacked to increase the depth of the model, as illustrated in Figure~\ref{fig:model_architecture}. 
Finally, the classifier module is composed of a $1\times1$ convolutional layer \cite{lin2013network} with $C$ filters followed by max pooling and a fully connected layer with $H$ hidden units and an output size $O$ equal to the number of possible answers (Gray in  Figure~\ref{fig:model_architecture}). A softmax layer predicts the probabilities of the answers. 
In order to use the Visual FiLM as a baseline for our experiments, we extract a 2D spectro-temporal representation of the acoustic scenes as depicted at the bottom of Figure~\ref{fig:model_architecture}.
The Resnet101 pre-trained extractor expects a 3 channels visual input but the spectro-temporal representation comprises only 1 channel.
To work around this constraint, the spectro-temporal information is simply repeated 3 times thus creating a 3 channels input (only when using Resnet101 as feature extractor).
This modified spectro-temporal representation is then fed to the model as if it was an image which is the simplest way to adapt the unmodified visual architecture to acoustic data.
We call this architecture \VisualFiLM.

\begin{figure*}
    \begin{center}
        \begin{subfigure}{.48\textwidth}
        \begin{center}
            \includegraphics[width=20em]{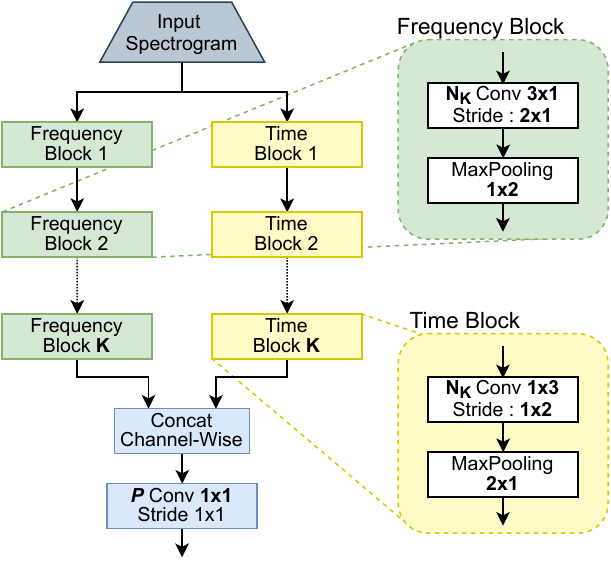}
            \caption{\textbf{Parallel feature extraction.} The input spectrogram is processed by 2 parallel pipelines. The first pipeline (in green) captures \emph{frequency} features using a serie of $K$ 1D convolutions with $N_k$ filters and a stride of $2\times1$. Since the stride is larger than $1\times1$, each convolution downsample the frequency axis. The $1\times2$ maxpooling then downsamples the time axis. The second pipeline (in yellow) captures \emph{time} features using the same structure with transposed filter size. Features from both pipelines are concatenated and fused using a $1\times1$ convolution with $P$ filters to create.}
            \label{fig:parallel_extractor}
        \end{center}
        \end{subfigure}
        \hspace{.03\textwidth}
        \begin{subfigure}{.48\textwidth}
        \begin{center}
            \includegraphics[width=17em]{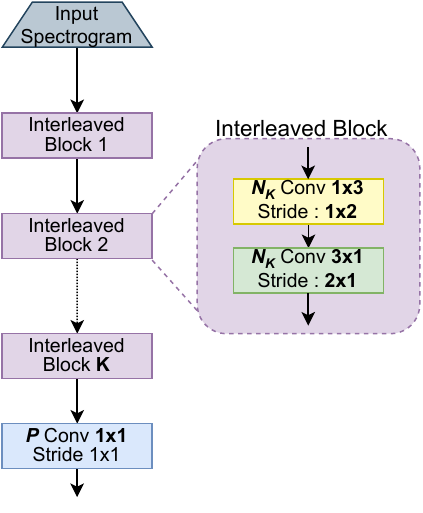}
            \caption{\textbf{Interleaved feature extraction.} 1D time (in yellow) and frequency (in green) convolutions are applied alternately on the input spectrogram building a \textbf{time-frequency} representation after each block. The order of the convolution in each block can be reversed. The extractor is composed of $K$ blocks where each convolution has $N_k$ filters followed by a $1\times1$ convolution with $P$ filters.}
            \label{fig:interlaced_extractor}
        \end{center}
      \end{subfigure}
      \caption{Acoustic feature extraction}
      \label{fig:feature_extraction}
  \end{center}
  \vspace{-1em}
\end{figure*}

\subsection{The proposed NAAQA architecture}
To create the NAAQA architecture, we made modifications to the baseline architecture that will be described in the following sections. The code is available on GitHub\footnote{\url{https://github.com/NECOTIS/NAAQA-Acoustic-Question-Answering}}.

\subsubsection{Feature Extraction}
\label{sec_method_feature_extraction}
As in Visual FiLM, the first step in the NAAQA model is feature extraction (orange box in Figure~\ref{fig:model_architecture}).
The most obvious adaptation of Visual FiLM to acoustic data is to retrain the feature extraction module on the scenes from CLEAR2.
To do this, we used three 2D convolutional layers, with $3\times3$ kernels, stride $2\times2$ and $N_1$, $N_2$, and $N_3$ filters respectively followed by a $1\times1$ convolutions with $N_4$ filters. We refer to this model as \NAAQATwoDConv.

However, as acoustic signals present unique properties, we introduce two feature extraction modules that are specifically tailored to sounds:
the \emph{Parallel} feature extractor (Figure~\ref{fig:feature_extraction}a) processes time and frequency features independently in parallel pipelines;
the \emph{Interleaved} feature extractor (Figure~\ref{fig:feature_extraction}b) captures time and frequency features in a single convolutional pipeline.
In both cases, the feature extractor is trained end-to-end with the rest of the network and uses a combination of 1D convolutional filters to process a 2D spectro-temporal representation.
The 1D filters process the time and frequency axis independently as opposed to the 2D filters typically used in image processing.

The design of the \emph{Parallel} feature extractor (Figure~\ref{fig:parallel_extractor}) is inspired by the work of \citeauthor{Pons2016}~\cite{Pons2016} where 1D filters are used to capture time and frequency features separately.
While \citeauthor{Pons2016} time-frequency model includes only 1 time and 1 frequency convolution which are then concatenated together, our extractor stacks multiple 1D convolutions in two parallel pipelines.
The time and frequency features are only fused at the end of both pipelines.
This yields more complex features.
The \emph{frequency pipeline} (green in the figure) comprises a serie of $K$ frequency blocks. Each block is composed of a 1D convolution with $N_K$ $3\times1$ kernels and $2\times1$ strides followed by a $1\times2$ maxpooling. 
With a stride larger than $1\times1$, the convolution operation downsamples the frequency axis and the pooling operation downsamples the time axis. 
This downsampling strategy allows features in both parallel pipelines to be of the same dimensions.
The \emph{time pipeline} (yellow in the figure) is the same as the frequency pipeline except that the convolutional kernel operates along the time dimension and the pooling along the frequency dimension. The convolution kernel is $1\times3$ and the pooling kernel $2\times1$. 
The activation maps of both pipelines are concatenated channel-wise and a representation combining both the time and frequency features is created using a $1\times1$ convolution \cite{lin2013network} with $P$ filters and a stride of one.
The feature maps dimensionality is either compressed or expanded depending on the number of filters $P$ in the $1\times1$ convolution. 
We name the corresponding model as \NAAQAParallel.
The $1\times1$ convolution can also be removed thus leaving it up to the next $3\times3$ convolution to fuse the time and frequency features.

The \emph{Interleaved} feature extractor (Figure~\ref{fig:interlaced_extractor}) processes the input spectrogram in a single pipeline composed of a serie of $K$ interleaved blocks (purple in the figure).
Each block comprises a $1\times3$ \emph{time} convolution with $N_K$ filters and stride $1\times2$ followed by a $3\times1$ \emph{frequency} convolution with $N_K$ filters and stride $2\times1$. 
A $1\times1$ convolution with $P$ filters processes the output of the last block to either compress or expand its dimensionality.
We name the corresponding model as \NAAQAInterleavedTime.

As an alternative configuration, the order of the convolution operation in each block can be reversed so that it first operates along the frequency axis and then the time axis.
The model is called \NAAQAInterleavedFreq, in this case.
Compared to the \emph{Parallel} feature extractor, time-frequency representations are created after each block instead of only at the end of the pipeline.

For all extractors, the convolutions in the first block comprise $N_1$ convolutional filters and the number of filters is doubled after each block ($N_i$ = $2N_{i - 1}$).
More blocks (higher $K$) gives a larger downsampling of the feature maps which brings down the computational cost of the model.



\subsubsection{Coordinate maps for acoustic inputs}
\label{method:coordconv}
When tackling the VQA task, the Visual FiLM model concatenates coordinate maps (\texttt{CoordConv}~\cite{liu2018intriguing}) to the input of convolutional layers (pink border boxes in Figure~\ref{fig:model_architecture}).
In the visual domain both axis of an image encode spatial information.
Coordinate maps have, therefore, the same meaning in the $x$ or $y$-axis.

In spectro-temporal representations for audio, however, the $y$-axis corresponds to frequency and the $x$-axis to time.
We, therefore, call the maps \emph{frequency} and \emph{time} coordinate map, respectively.
All spectro-temporal representations in CLEAR2 have the same range for the frequency axis but the range for the time axis varies depending on the duration of the acoustic scenes. 
We hypothesize that \emph{time} coordinate maps might have a stronger impact on performance because they provide a relative time scale that the model can use to enhance its temporal localization capabilities.

\subsubsection{Complexity Optimization}
\label{sec:method:complexity}
We performed optimization of the most important hyper-parameters in the NAAQA architecture with the goal of reducing model complexity.
These include number of \emph{GRU} text-processing units $G$; the number of Resblock $J$ that dictates the number of FiLM layers and, therefore, the number of modulation coefficients to compute; the number of convolutional filters $M$ in each block; the number of filters $C$ and the number of hidden units $H$ in the classifier module.
We refer to the resulting model by prepending \textbf{\texttt{Optimized}} to the model name.

\subsubsection{NAAQA with a MALiMo module}
In MALiMo~\cite{Fayek2019}, \citeauthor{Fayek2019} add a second set of FiLM layers that acts has an auxiliary controller.
The controller uses the extracted acoustic features to further modulate the intermediate Resblocks.
To evaluate the impact of MALiMo on CLEAR2, a MALiMo module was added to NAAQA.
We refer to this configuration by appending \MALiMo to the names introduced above.
In our implementation of the module we replaced LSTMs with GRUs and adapted the inputs to the acoustic features that we study.

\section{Experiments}
\label{sec:experiments}
We perform experiments to compare the effect on performance of our modification to the baseline model.
Most experiments are conducted on the proposed CLEAR2 dataset.
We first investigate 
different feature extraction methods and compare them to the \VisualFiLM feature extractor.
Then, we show the effect of time and frequency coordinate maps at different levels of the model.
Moreover, we perform an hyper-parameters ablation study to reduce the complexity of the model.
We finally test the addition of a MALiMo module to our model.
To demonstrate the generality of the results, we compare the performance of our model on CLEAR2 and \DAQAp datasets.

\subsection{Acoustic Pre-processing}
\label{sec:preprocessing}

The raw acoustic signal (sampled at 48 kHz for CLEAR2 and 16kHz for \DAQAp) is processed to create a 2D time-frequency representation with Mel scale~\cite{stevens1937scale} spectrograms.
After preliminary tests it was decided to extract 64 Mel coefficients for both CLEAR2 and \DAQAp computed over samples weighted by a Hanning window.
The window size was of 512 samples ($\sim$10.6 msec) for CLEAR2 whereas for \DAQAp it was of 400 samples  ($\sim$25ms) as in~\cite{Fayek2019}.
The time shift between consecutive windows (stride) was also optimized depending on the characteristics of audio data.
We found that the best results for CLEAR2 was a time shift of 2048 samples ($\sim$42.7 msec).
This is feasible because 
of the sustained notes 
which vary slowly in time.
Using such a long time shift allowed us to reduce more than ten folds the computational costs. 
As \DAQAp contains sounds that are shorter and less stable, the same optimization is not feasible.
In fact, with a time shift of 1600 samples (100ms) a 5\% drop in accuracy is observed in comparison with 160 samples (10ms) shifts.
All results based on CLEAR2 are reported with long window shifts (long stride), with the exception of the comparison between short and long strides on both CLEAR2 and \DAQAp in Supplementary Materials.

As duration of scenes are not constant in CLEAR2, spectrograms are zero padded along the time axis 
so that they all have the same dimension ($1 \times 64 \times 418$) which corresponds to a maximum length of $\sim$17.9 sec.
The power spectrum is normalized to the mean and standard deviation of the training data with the goal of
speeding up convergence~\cite{lecun2012efficient}.

\subsection{Experimental conditions}
\label{exp:exp_conditions}
Unless specified otherwise, the models presented in subsequent sections are trained on the CLEAR2 dataset which comprises 50 000 scenes and 4 questions per scene for a total of 200 000 records from which 140 000 (70\%) are used for training, 30 000 (15\%) for validation and 30 000 (15\%) for test. 
The test set is generated using a different set of elementary sounds which ensures that the network could not memorize them and can therefore acts has a better generalization benchmark. The optimization techniques and other training settings are further described in Supplementary Materials.
Results are reported in terms of accuracy, that is in percentage of correct answers over the total. 
Since initialization of deep architectures has a profound impact on training convergence, we developed a python library \texttt{torch-reproducible-block}\footnote{\url{https://github.com/NECOTIS/torch-reproducible-block}}
to control the model initial conditions and design reproducible experiments.
To ensure the robustness of the results, each model is trained 5 times with 5 different random seeds.

\subsection{Initial model configuration}
\label{exp:initial_config}
The initial configuration for the proposed model comprises $G=4096$ GRU units, $J=4$ Resblocks with $M=128$ filters each and a classifier composed of a $1\times1$ convolution with $C=512$ filters and $H=1024$ hidden units in the fully connected layer.
This configuration includes both \emph{time} and \emph{frequency} coordinate maps in each location highlighted in pink in Figure \ref{fig:model_architecture}.

\begin{table*}
    \centering
    \resizebox{\textwidth}{!}{
        \addtolength{\tabcolsep}{-3pt}    
        \begin{tabular}{lccccccccccccc}
            \hline
            \hline
			 			         & Number of & Overall & \multicolumn{11}{c}{Accuracy by question type (\%)} \\
			 Configuration & Parameters  & Acc. & Instrument & Note & Brightness & Loudness & Exist & Abs. Pos. & Glob. Pos. & Rel. Pos. & Count & Count Comp. & Count Inst.  \\
            \hline
            \multicolumn{14}{l}{Baselines} \\
            \hline
    Random Answer        & --- & 1.75 & 1.75 & 1.75 & 1.75 & 1.75 & 1.75 & 1.75 & 1.75 & 1.75 & 1.75 & 1.75 & 1.75 \\
    Most common Answer (Yes)   & --- & 7.3 & 0 & 0 & 0 & 0 & 55.62 & 0 & 0 & 0 & 0 & 47.27 & 0 \\
    \VisualFiLM & 6.71 M & 62.3 ±0.78 & 61.9 ±0.85 & 37.8 ±1.30 & 83.9 ±0.58 & 81.9 ±0.61 & 73.2 ±0.70 & 51.7 ±1.40 & 73.1 ±2.29 & 48.4 ±1.82 & 41.5 ±0.42 & 59.8 ±0.57 & 39.9 ±3.03 \\
             \hline
             \multicolumn{14}{l}{NAAQA} \\
             \hline
    \NAAQATwoDConv & 5.61 M & 77.6 ±0.72 & 80.6 ±0.66 & 73.4 ±1.28 & 90.5 ±1.44 & 86.5 ±0.76 & 81.3 ±0.57 & 74.7 ±2.50 & 87.5 ±0.42 & 54.0 ±1.43 & 53.8 ±1.01 & 60.8 ±1.34 & 49.0 ±3.53 \\
    \NAAQAInterleavedFreq & 5.61 M & 67.2 ±0.98 & 63.0 ±1.93 & 48.0 ±2.14 & 84.5 ±1.07 & 81.1 ±0.52 & 72.1 ±1.24 & 65.1 ±1.04 & 80.1 ±0.66 & 49.3 ±1.59 & 42.5 ±1.51 & 56.6 ±2.45 & 46.5 ±1.57 \\
    \NAAQAInterleavedTime & 5.61 M & 78.0 ±0.51 & \textbf{81.8 ±1.06} & 70.6 ±1.24 & 90.9 ±0.78 & \textbf{87.9 ±0.94} & \textbf{81.6 ±0.66} & 75.9 ±1.08 & 87.4 ±0.25 & \textbf{60.0 ±4.07} & 53.6 ±0.63 & \textbf{61.2 ±0.73} & 50.0 ±2.23 \\
    \NAAQAParallel & 5.61 M & 78.5 ±0.45 & 80.4 ±0.83 & 72.6 ±1.19 & 91.2 ±0.74 & 87.1 ±0.28 & 80.4 ±0.16 & 78.7 ±1.26 & 88.8 ±0.53 & 55.4 ±1.63 & 52.6 ±0.43 & 59.8 ±2.13 & 50.2 ±3.49 \\
    \NAAQAParallelOptimized & 1.68 M & \textbf{79.5 ±0.05} & 81.7 ±0.20 & \textbf{74.2 ±0.58} & \textbf{91.9 ±0.17} & 87.4 ±0.31 & 81.2 ±0.10 & \textbf{79.3 ±0.77} & \textbf{90.0 ±0.15} & 58.0 ±1.32 & \textbf{53.8 ±1.40} & 60.6 ±0.89 & \textbf{50.4 ±1.10} \\

             \hline
             \multicolumn{14}{l}{NAAQA + MALiMo} \\
             \hline
    \VisualFiLM + \MALiMo & \textbf{6.82 M} & 63.6 ±1.20 & 61.8 ±1.40 & 36.3 ±1.41 & 83.5 ±0.86 & 82.3 ±0.59 & 73.4 ±1.06 & 57.1 ±3.58 & 76.1 ±1.94 & 47.4 ±2.98 & 41.6 ±0.80 & 56.7 ±4.23 & 43.2 ±3.64 \\
    \NAAQATwoDConv + \MALiMo & 6.71 M & 77.1 ±1.12 & 79.2 ±1.01 & 71.3 ±1.54 & 90.5 ±0.81 & 86.1 ±0.73 & 80.3 ±0.77 & 75.7 ±2.08 & 87.8 ±0.74 & 52.7 ±1.94 & 51.9 ±2.73 & 59.0 ±3.00 & 50.2 ±2.48 \\
    \NAAQAInterleavedFreq + \MALiMo & 6.71 M & 64.8 ±3.91 & 55.3 ±8.33 & 43.1 ±8.77 & 83.2 ±1.33 & 80.0 ±1.75 & 70.3 ±2.74 & 63.6 ±4.72 & 78.7 ±3.27 & 45.8 ±3.53 & 41.4 ±1.96 & 55.1 ±2.75 & 48.7 ±2.05 \\
    \NAAQAInterleavedTime + \MALiMo & 6.71 M & 77.1 ±0.63 & 79.9 ±0.86 & 70.8 ±1.10 & 90.5 ±1.00 & 86.7 ±0.89 & 80.3 ±0.34 & 74.5 ±1.06 & 87.6 ±0.35 & 55.2 ±2.39 & 53.8 ±0.43 & 59.9 ±2.32 & 50.0 ±0.93 \\
    \NAAQAParallel + \MALiMo & 6.71 M & 77.3 ±0.93 & 78.2 ±1.50 & 71.3 ±1.00 & 89.9 ±0.66 & 85.8 ±0.67 & 79.7 ±0.45 & 77.7 ±2.02 & 87.6 ±0.53 & 53.6 ±1.84 & 51.9 ±1.82 & 59.0 ±2.29 & 48.1 ±1.65 \\
    \NAAQAParallelOptimized + \MALiMo & 2.78 M & 78.2 ±0.06 & 80.8 ±0.41 & 72.4 ±0.33 & 89.6 ±0.62 & 86.2 ±0.04 & 79.7 ±0.10 & 79.3 ±0.22 & 88.4 ±0.36 & 54.0 ±0.33 & 52.2 ±0.70 & 58.8 ±0.45 & 45.6 ±0.00 \\
             \hline
             \hline
        \end{tabular}
    }
    \caption{\textbf{Results on CLEAR2.} Table gives the number of parameters, average accuracy (\%), and standard deviation over 5 repetitions of the training. Overall accuracy as well as question-kind dependent accuracy are reported. Different configurations are reported in the same order as they are discussed in the paper. The most common answer is ``Yes''.}
    \label{tab:results:clear2}
\end{table*}

\begin{table*}
    \centering
        \begin{tabular}{lcccccr}
            \hline
            \hline
            Configuration   & \# Parameters & Train Acc. & Val Acc. & Test Acc. & Trainig time \\
            \hline
    \NAAQATwoDConvOptimized & 1.68 M & 65.2 ±0.64 & 58.0 ±0.82 & 58.3 ±0.98 & 0 days 06:48:12 \\
    \NAAQAParallelOptimized & 1.68 M & 66.6 ±0.51 & 60.4 ±0.08 & 60.4 ±0.21 & 0 days 06:49:52 \\
            \hline
    \NAAQATwoDConvOptimized + \MALiMo  & \textbf{2.78 M} & 58.5 ±3.03 & 54.2 ±2.42 & 54.4 ±2.35 & 0 days 07:31:37 \\
    \NAAQAParallelOptimized + \MALiMo & \textbf{2.78 M} & \textbf{67.3 ±1.22} & \textbf{64.1 ±0.54} & \textbf{64.3 ±0.72} & \textbf{1 days 04:59:46} \\
        \hline
        \hline
        \end{tabular}
    \caption{\textbf{Results on \DAQAp} The table presents number of parameters, average training, validation and test accuracy (\%) with standard deviation over 5 repetitions of the training as well as average training time. Results are reported for four configurations, with and without the MALiMo module in the same order as they are presented in the paper.}
    \label{tab:q_type:daqa}
\end{table*}

\section{Results and Discussion}
\label{sec:results_and_discussion}
Main results on the CLEAR2 data set are presented in Table~\ref{tab:results:clear2}.
The complexity of the models in terms of number of parameters, the overall accuracy and accuracy dependent on question's type are reported.
Results from two theoretical baselines  - random chance and majority class answers - are first given.
Then we report results from the  \VisualFiLM baseline model with the initial configuration described in Section \ref{exp:initial_config}. This architecture achieves the lowest accuracy of 62.3\% in comparison with all tested models.
As expected, the pre-learned knowledge gathered in a visual context does not transfer directly to the acoustic context. Mel spectrograms have a very different structure than visual scenes features. 

\subsection{NAAQA modifications}

Unless specified otherwise, the initial configuration described in section \ref{exp:initial_config} is used for all models in this section.

\subsubsection{Feature Extraction}
\label{exp:feat_extractor}
The first improvement to the baseline is given by introducing a specific audio feature extraction module based on 2D convolutions. The \NAAQATwoDConv model has slightly fewer parameters than \VisualFiLM because of the simpler feature extraction module and a much higher overall accuracy of 77.6\%.



We then tested two versions of the \emph{Interleaved} feature extractor (Figure~\ref{fig:interlaced_extractor}).
The computation order of the 1D convolutions in each block has a significant impact on performance. When the first 1D convolution in each block is computed along the frequency axis (\NAAQAInterleavedFreq), the network reaches an overall accuracy of 67.2\%.
It performs especially poorly with questions related to \emph{count} (42.5\%), \emph{count instruments} (46.5\%) and \emph{notes} (48.0\%).
The performance on \emph{position} questions is also the lowest among all  extractors.
When the computation order of the convolution is reversed (\NAAQAInterleavedTime), information is better captured and the network reaches 78.0\% of overall accuracy.
A possible explanation relates to the nature of the sounds in the CLEAR2 dataset which mainly consists of sustained musical notes.
The time dimension at short scales does not contain much information that helps identifying the individual sounds.
At larger scales, however, the time axis contains information relative to the scene as a whole which is exploited by higher level layers (Resblocks) to take into account the connections between different sounds.
Because its stride is greater than 1x1, each 1D convolution downsamples the axis on which it is applied.
When the first 
is a frequency convolution, the frequency axis of the resulting features is downsampled which reduces the information that can be captured by the time convolution that follows.

The \emph{Parallel} feature extractor (\NAAQAParallel, Figure \ref{fig:parallel_extractor}) reaches an overall accuracy of 78.5\%.
It performs well on all question's types except \emph{relative position}, \emph{count} and \emph{count instrument}.
Refer to Section \ref{exp:general_clear} for further analysis.
These results show that building complex time and frequency feature separately and fusing them at a later stage is a good strategy to learn acoustic features for this task.
This claim is further strengthen by the analysis of section \ref{exp:NAAQA_MALIMO_on_DAQA}.

Out of all extractors, \NAAQAParallel is the one that performs the best and constitutes the basis of NAAQA in all subsequent experiments.


\subsubsection{Coordinate Maps}
\label{exp:coord_conv}

\begin{table}
    \centering
    \resizebox{\columnwidth}{!}{
        \begin{tabular}{ccccccc}
            \hline
            \hline
            \multicolumn{4}{c}{Coordinate maps} & \multicolumn{3}{c}{Accuracy (\%)} \\
            Extractor & 1st Conv & Resblocks & Classifier & Train & val & Test \\
            \hline
  -- & -- & Time & -- & 95.0 ±0.79 & \textbf{90.2 ±0.62} & \textbf{79.0 ±0.44} \\
  -- & Time & -- & -- & \textbf{95.1} ±0.90 & 90.0 ±0.28 & \textbf{79.0 ±0.43} \\
  -- & -- & Both & -- & 94.9 ±1.08 & 90.1 ±0.83 & 78.8 ±0.52 \\
  -- & Both & -- & -- & 95.1 ±1.23 & 90.0 ±0.46 & 78.7 ±0.70 \\
Time & -- & -- & -- & 94.7 ±1.06 & 88.3 ±1.24 & 76.7 ±0.83 \\
Both & -- & -- & -- & 94.1 ±1.47 & 88.4 ±1.06 & 76.6 ±0.36 \\
  -- & -- & -- & Freq & 84.0 ±2.17 & 73.5 ±1.40 & 64.8 ±1.51 \\
  -- & -- & -- & -- & 85.2 ±0.71 & 72.9 ±1.36 & 63.8 ±0.70 \\
  -- & -- & -- & Both & 83.5 ±3.46 & 72.3 ±2.71 & 63.5 ±1.93 \\
  -- & -- & Freq & -- & 84.5 ±0.67 & 71.7 ±2.85 & 62.6 ±2.36 \\
  -- & Freq & -- & -- & 83.4 ±1.53 & 70.7 ±1.74 & 61.7 ±1.16 \\
  -- & -- & -- & Time & 81.9 ±0.88 & 70.1 ±1.99 & 61.7 ±0.76 \\
Freq & -- & -- & -- & 79.7 ±3.70 & 67.1 ±3.36 & 59.3 ±2.20 \\
            \hline
            \hline
        \end{tabular}
    }
    \caption{\textbf{Impact of the placement of \emph{Time} and \emph{Frequency coordinate maps}.} All possible positions are illustrated by the pink border boxes in Figure \ref{fig:model_architecture}. The value \emph{Both} indicate that both \emph{Time} and \emph{Frequency} coordinate maps were inserted at the given position. The \NAAQAParallel is used with hyper-parameters from the initial configuration (defined in section \ref{exp:exp_conditions}. The rows are ordered by test accuracy.). 
    }
    \label{tab:coordconv}
\end{table}

Coordinate maps can be inserted before any convolution operation (Figure~\ref{fig:model_architecture}).
We therefore analyzed the impact of the placement of \emph{Time} and \emph{Frequency} coordinate maps at different depths in the network.
All possible locations were evaluated via grid-search.
For each location, we inserted either a \emph{Time} coordinate map, a \emph{Frequency} coordinate map or both.
Results are detailed in Table \ref{tab:coordconv}.
\emph{Time} coordinate maps have the biggest impact on performance, especially when inserted in the first convolution after the feature extractor or in the Resblocks.
This could be because the fusion of the textual and acoustic features, and therefore most of the reasoning, is performed in the Resblocks.
The network might be using the additional localization information to inform the modulation of the convolutional feature maps based on the context of the question.
Surprisingly, the \emph{Frequency} coordinate maps have a minimal impact on performance.
We further compare the impact of \emph{Time} versus \emph{Frequency} coordinate maps in Supplementary Materials.

\subsubsection{Complexity Optimization}
\label{sec:exp_baseline_optimization}

\begin{figure}
  \includegraphics[width=\columnwidth]{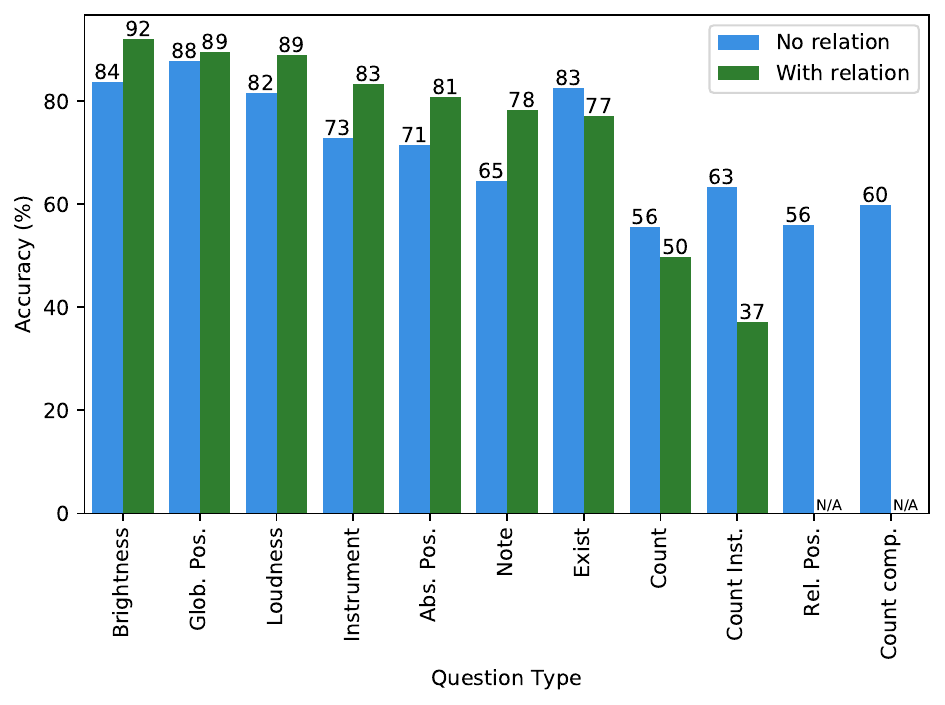}
  \caption{\textbf{Test accuracy by question type and the number of relation in the question for \NAAQAParallelOptimized.} The overall accuracy for this configuration is 79.1\%. The presence of \emph{before} or \emph{after} in a question constitutes a temporal relation.
  The accuracy is \emph{N/A} for \emph{relative position} and \emph{count compare} since these types of question do no include relations. The hyper-parameters are described in the end of Section \ref{sec:exp_baseline_optimization}}
  \label{fig:naaqa_per_q_type}
\end{figure}

As described in Section~\ref{sec:method:complexity}, we optimized the most important hyper-parameters ($G,J,M$) in the NAAQA model to reduce its complexity.
The baseline \VisualFiLM configuration comprises 6.71 M parameters and achieves only 62.3\%.
\NAAQAParallel comprises 5.61 M parameters and performs significantly better with 78.5\%.
With this model as a starting point, we performed an ablation study to find which hyper-parameters can be reduced without impacting accuracy.
The \NAAQAParallelOptimized configuration is the best trade-off between model complexity and performance.
It comprises 1.68 M parameters and achieves the best overall accuracy with 79.5\%.
The most notable complexity reduction comes from the reduction of the number of GRU units $G$. 
Reducing $G$ from $4096$ to $512$ increased accuracy while reducing the number of parameters by a factor of 3 (6.61 M vs 1.68 M).
The \NAAQAParallelOptimized is composed of a \emph{Parallel} extractor with $K=3$ blocks and $P=64$, $G=512$ GRU units, $J=4$ Resblocks with $M=128$ filters, a classifier module with $C=512$ filters and $H=1024$ units.
Results for this configuration can be found in Table \ref{tab:results:clear2} and Figure \ref{fig:naaqa_per_q_type}.
Further results related to the ablation study can be found in Supplementary Materials.

\subsubsection{Adding a MALiMo controller}
\label{exp:adding_malimo_module}
The bottom rows of Table~\ref{tab:results:clear2} show results where the configurations described in previous sections are augmented with a MALiMo controller.
Although the model complexity is significantly increased ($\sim$1M parameters), this addition does not bring any improvement in the model performance on CLEAR2.
Almost all the tested configurations with a MALiMo controller perform slightly worse than the same configuration without the module, as can be seen in Table~\ref{tab:results:clear2}.
This may be again explained by the characteristics of the sounds in CLEAR2.
A more in-depth discussion is given when we evaluate the models on \DAQAp.

\vspace{-2mm}
\subsection{Summary of Results on CLEAR2}
\label{exp:general_clear}
NAAQA performs well on the CLEAR2 AQA task with 79.5\% overall accuracy.
It does however struggle with certain types of question as shown in Table~\ref{tab:results:clear2} and Figure~\ref{fig:naaqa_per_q_type}.
When asked to \emph{count} the number of sounds with specific attributes, NAAQA reaches only 53.8\% accuracy.
This limitation is more severe if the question is to count the \emph{different instruments} playing in a given part of the scene (50.4\%).
It attains slightly higher accuracy when asked to compare the number of instances of acoustic objects (more, fewer or equal number) with specific attributes (60.6\%).
In contrast, the network can successfully recognize individual instruments in the scene (81.7\%).
This suggests, that the problem lies in the logical complexity of the question rather than in the pattern matching from the acoustic scene.
As an example, the question (count instrument): "How many different instruments are playing after the third cello playing a C\# note?" requires to first identify the cello playing the \emph{C\#} note, then identify all acoustic objects that are playing after this sound, determine which instruments are of the same family and finally count the number of different families. The model struggles when it must focus on a large number of acoustic objects which explains the low accuracy for this type of question.

A similar argument could explain why models also have difficulties with questions related to the \emph{relative position} of the instruments (58.0\%).
For example, to answer the question "Among the flute sounds, which one plays an F note?", the model must find all flutes playing in the scene, determine which one plays an F note, counts the number of flute playing before and translates the count to a position~\footnote{This is one possible strategy to answer the question. There may be other ways.}.
This also requires the network to focus on multiple objects.

Certain questions include temporal relations between sounds (\emph{before} and \emph{after}) as exemplified in Table \ref{tab:types_of_question}.
Questions that include relations require focusing on several sounds to be answered.
Figure~\ref{fig:naaqa_per_q_type} shows the accuracy for each question type depending on the presence of temporal relations.
Questions that require the network to focus on a single acoustic object (\emph{brightness}, \emph{loudness}, \emph{instrument}, \emph{note}, \emph{global position} and \emph{absolute position}) benefit from the presence of a relation in the question.
This might be explained by the fact that the question contains more information about the scene which helps to focus on the right acoustic object.
However, the presence of relations in questions that already require the network to focus on multiple objects (\emph{exist}, \emph{count} and \emph{count comparison}) is detrimental.
This again supports the idea that having to focus on too many objects in the scene hinders the network performance.

\subsection{Evaluation on \DAQAp}
To compare our results to those of \citeauthor{Fayek2019}~\cite{Fayek2019}, we evaluated our models on a version of the DAQA data set.
As mentioned in Section \ref{data:reproducing_daqa}, we were not able to reproduce the original DAQA dataset which means that results presented in this section are not fully comparable with \cite{Fayek2019}.
Results for different configurations of NAAQA tested on our modified \DAQAp are reported in Table \ref{tab:q_type:daqa}.

\subsubsection{NAAQA on \DAQAp}
\label{exp:NAAQA_on_DAQA}
The models explored in this section matches the performance of previous efforts~\cite{Fayek2019}. 
The smallest model they evaluated had 5.49M parameters, the biggest model had 21.33M parameters and the best performing model had 13.20M parameters.
The \NAAQATwoDConvOptimized model only has 1.68 M parameters and reaches an accuracy of 58.3\% on \DAQAp.
The \NAAQAParallelOptimized has the same number of parameters and performs slightly better with and accuracy of 60.4\%.
When we analyzed both of these models on CLEAR2 dataset in Section \ref{exp:feat_extractor}, we found a much smaller difference between the performance of the \NAAQATwoDConv and the \NAAQAParallel.
This difference suggests that the parallel extractor is more effective in the context of complex acoustic sounds (\DAQAp) than with sustained musical notes (CLEAR2).

Even though these results are not 100\% comparable with \cite{Fayek2019} because of the difference in the dataset composition, we want to emphasize that \NAAQAParallelOptimized reaches a somewhat similar accuracy than the smallest FiLM in \cite{Fayek2019} (60.4\% vs 64.3\%) with significantly smaller number of parameters (1.68 M vs 5.49 M).

\subsubsection{NAAQA with a MALiMo module on \DAQAp}
\label{exp:NAAQA_MALIMO_on_DAQA}
In Section \ref{exp:adding_malimo_module}, we found that adding a MALiMo controller to our NAAQA models did not improve the accuracy on CLEAR2.
On the other hand,  the \mbox{MALiMo} controller has a significant positive impact when the model is evaluated on \DAQAp dataset (Table \ref{tab:q_type:daqa}). 
We see an increase of almost 4\% when using \NAAQAParallelOptimized + \MALiMo compared to \NAAQAParallelOptimized alone.
These results are consistent with \citeauthor{Fayek2019} findings and with the hypothesis that MALiMo increases performance when working with complex sounds.


The \NAAQAParallelOptimized + \MALiMo configuration performs about the same as the smallest MALiMo model evaluated in \cite{Fayek2019} (64.3\% vs 65.1\%) with significantly fewer parameters (2.78 M vs 8.91 M).

\vspace{-3mm}
\section{Conclusions}
\label{sec:conclusions}
Acoustic Question Answering (AQA) is a newly emerging task in the area of machine learning research.
As performance is strongly dependent on the acoustical environments and types of questions, it is important to understand the relationship between the application and the chosen neural architecture.
We propose a benchmark for AQA based on musical sounds (CLEAR2) and a neural architecture that is tailored to interpreting acoustic scenes (NAAQA).
NAAQA introduces a number of modifications to a FiLM based architecture to optimize acoustic scenes analysis.
These includes several strategies for neural feature extraction, an ablation study of the hyper-parameters and the optimization of coordinate maps. 
We confirm that FiLM layers are very effective to modulate activation maps in the AQA application.
We are able to optimize our NAAQA neural network so to obtain competitive results with a fraction of the model complexity.
These results are confirmed on a different AQA task (\DAQAp) comprising more complex sounds with the addition of a MALiMo controller in the model.
We release all code openly in the hope that these resources may foster increased research activity in solving the AQA task.


\vspace{-3mm}

\section*{Acknowledgment}
We would like to thank the reviewers for their constructive comments that helped us improve the paper. The NVIDIA Corporation for the donation of GPUs. Part of this research was funded or supported by the CHIST-ERA IGLU project, the CRSNG, the Michael-Smith scholarships, the Universities of Sherbrooke and NTNU.

\printbibliography

@string{TREC="TREC"}

@string{SIGIR="SIGIR"}

@string{EMNLP="EMNLP"}

@string{ACL="ACL"}

@string{CVPR="CVPR"}

@string{ICCV="ICCV"}

@string{NeurIPS="NeurIPS"}

@string{NeurIPSVigil="NeurIPS Vigil Workshop"}

@string{PNAS="PNAS"}

@string{HICSS="HICSS"}

@string{ISPA="ISPA"}

@string{MM="MM"}

@string{IJCAI="IJCAI"}

@string{IEEETCSVT="IEEE T CIRC SYST VID"}

@string{Diagrams="Diagrams"}

@string{IROS="IROS"}

@string{IEEEAUDIO="IEEE-ACM T AUDIO SPE"}

@string{AAAI="AAAI"}

@string{ASIST="ASIS\&T"}

@string{ICLR="ICLR"}

@string{ICML="ICML"}

@string{ECCV="ECCV"}

@string{EUSIPCO="EUSIPCO"}

@string{ICASSP="ICASSP"}

@string{ISMIR="ISMIR"}

@string{CBMI="CBMI"}

@string{IJCV="IJCV"}

@string{DCASE="DCASE"}

@string{JASA="JASA"}

@string{JMLR="JMLR"}

@string{IEEETPAMI="TPAMI"}

@inproceedings{audioset,
title	= {Audio {S}et: {A}n ontology and human-labeled dataset for audio events},
author	= {{G}emmeke, Jort {F}. and {E}llis, {D}aniel {P}. {W}. and {F}reedman, {D}ylan and {J}ansen, {A}ren and {L}awrence, {W}ade  and {M}oore, {R}. {C}hanning and {P}lakal, {M}anoj and {R}itter, {M}arvin},
year	= {2017},
booktitle	= ICASSP
}

@inproceedings{pons2017timbre,
    title = {Timbre {A}nalysis of {M}usic {A}udio {S}ignals with {C}onvolutional {N}eural {N}etworks},
    author = {Pons, {J}ordi and {S}lizovskaia, {O}lga and {G}ong, {R}ong and G{\'{o}}mez, {E}milia and {S}erra, {X}avier},
    booktitle = EUSIPCO,
    pages     = {2744--2748},
    year = 2017,
}

@article{audioretrieval2022,
  title={{A}udio {R}etrieval with {N}atural {L}anguage {Q}ueries: {A} {B}enchmark {S}tudy}, 
  author={Koepke, {A}. {S}ophia and {O}ncescu, {A}ndreea-{M}aria and {H}enriques, {J}oao and {A}kata, {Z}eynep and {A}lbanie, {S}amuel},
  journal={{IEEE} {T}ransactions on {M}ultimedia}, 
  year={2022}
}

@inproceedings{brousmiche2020soundClassification,
  author={{Brousmiche}, Mathilde and {J}ean {Rouat} and {S}tephane {Dupont}},
  booktitle=ICASSP,
  title={{SECL-UMons} {D}atabase for {S}ound {E}vent {C}lassification and {L}ocalization}, 
  year={2020},
  pages={756--760},
  doi={10.1109/ICASSP40776.2020.9053298}}

@inproceedings{farnet2019,
   title={{Receptive-Field-Regularized} {CNN} {V}ariants for {A}coustic {S}cene {C}lassification},
   ISBN={9780578595962},
   url={http://dx.doi.org/10.33682/cjd9-kc43},
   DOI={10.33682/cjd9-kc43},
   booktitle=DCASE,
   author={Koutini, {K}haled and {E}ghbal-zadeh, {H}amid and {W}idmer, {G}erhard},
   year={2019}
}

@article{timbral_model,
  title={Timbral {M}odels, {A}udio{C}ommons project, {D}eliverable {D5.7}},
  author={Pearce, {A}ndy and {B}rookes, {T}im and {M}ason, {R}ussell},
  year={2018},
  url={http://www.audiocommons.org/materials/}
}

@inproceedings{liu2018intriguing,
  title={An intriguing failing of convolutional neural networks and the coordconv solution},
  author={Liu, {R}osanne and {L}ehman, {J}oel and {M}olino, {P}iero and {S}uch, {F}elipe {P}etroski and {F}rank, {E}ric and {S}ergeev, {A}lex and {Y}osinski, {J}ason},
  booktitle=NeurIPS,
  pages={9605--9616},
  year={2018}
}

@techreport{ITULoudness2015,
    title = {Algorithms to measure audio programme loudness and true-peak audio level ({ITU-R BS.1770-4})},
    author = {International {T}elecommunication {U}nion},
    year = {2015},
    institution = {Electronic {P}ublication},
    pages = {25},
    url = {https://www.itu.int/dms_pubrec/itu-r/rec/bs/R-REC-BS.1770-4-201510-I!!PDF-E.pdf}
}

@inproceedings{lin2013network,
    title = {Network {I}n {N}etworks},
    author = {Min {L}in and {Q}iang {C}hen and {S}huicheng {Y}an},
    year = {2014},
    booktitle=ICLR,
}

@inproceedings{das_bias,
    author = {Das, {A}nubrata and {A}njum, {S}amreen and {G}urari, {D}anna},
    title = {Dataset bias: A case study for visual question answering},
    booktitle = ASIST,
    pages = {58--67},
    doi = {10.1002/pra2.7},
    year = {2019}
}

@InProceedings{Hu_2019_ICCV,
    author = {Hu, {R}onghang and {R}ohrbach, {A}nna and {D}arrell, {T}revor and {S}aenko, {K}ate},
    title = {Language-Conditioned {G}raph {N}etworks for {R}elational {R}easoning},
    booktitle = ICCV,
    year = {2019},
    pages     = {10293--10302},
}

@InProceedings{hu_stack_clevr_2018,
    author = {Hu, {R}onghang and {A}ndreas, {J}acob and {D}arrell, {T}revor and {S}aenko, {K}ate},
    title = {Explainable {N}eural {C}omputation via {S}tack {N}eural {M}odule {N}etworks},
    booktitle = ECCV,
    year = "2018",
    pages = "55--71"
}

@InProceedings{prob_clevr2019,
    title = {Probabilistic {N}eural {S}ymbolic {M}odels for {I}nterpretable {V}isual {Q}uestion {A}nswering},
    author = {Vedantam, {R}amakrishna and {D}esai, {K}aran and {L}ee, {S}tefan and {R}ohrbach, {M}arcus and {B}atra, {D}hruv and {P}arikh, {D}evi},
    booktitle = ICML,
    pages = {6428--6437},
    year = {2019}
}

@InProceedings{Manjunatha_2019_CVPR,
    author = {Manjunatha, {V}arun and {S}aini, {N}irat and {D}avis, {Larry S.}},
    title = {Explicit {B}ias {D}iscovery in {V}isual {Q}uestion {A}nswering {M}odels},
    booktitle = CVPR,
    year = {2019},
    pages = {9562--9571},
}

@InProceedings{Hudson_2019_CVPR,
    author = {Hudson, {Drew A.} and {M}anning, {Christopher D.}},
    title = {{GQA}: A {N}ew {D}ataset for {R}eal-World {V}isual {R}easoning and {C}ompositional {Q}uestion {A}nswering},
    booktitle = CVPR,
    year = {2019},
    pages     = {6700--6709},
}

@InProceedings{Gordon_2018_CVPR,
    author = {Gordon, {D}aniel and {K}embhavi, {A}niruddha and {R}astegari, {M}ohammad and {R}edmon, {J}oseph and {F}ox, {D}ieter and {F}arhadi, {A}li},
    title = {{IQA}: {V}isual {Q}uestion {A}nswering in {I}nteractive {E}nvironments},
    booktitle = CVPR,
    year = {2018},
    pages     = {4089--4098},
}

@ARTICLE{wang_FVQA,
    author = {Wang, {P}eng and {W}u, {Q}i and {S}hen, {C}hunhua and {D}ick, {A}nthony and {Van {D}en {H}engel}, {A}nton},
    journal = IEEETPAMI,
    title = {{FVQA}: {F}act-{B}ased {V}isual {Q}uestion {A}nswering},
    year = {2018},
    pages = {2413--2427}
}

@inproceedings{NIPS2018_YI,
    title = {Neural-Symbolic {VQA}: {D}isentangling {R}easoning from {V}ision and {L}anguage {U}nderstanding},
    author = {Yi, {K}exin and {W}u, {J}iajun and {G}an, {C}huang and {T}orralba, {A}ntonio and {K}ohli, {P}ushmeet and {T}enenbaum, {Joshua B.}},
    booktitle = NeurIPS,
    pages = {1039--1050},
    year = {2018}
}

@InProceedings{Agrawal_2018_CVPR,
    author = {Agrawal, {A}ishwarya and {B}atra, {D}hruv and {P}arikh, {D}evi and {K}embhavi, {A}niruddha},
    title = {Don't {J}ust {A}ssume; {L}ook and {A}nswer: {O}vercoming {P}riors for {V}isual {Q}uestion {A}nswering},
    booktitle = CVPR,
    year = {2018},
    pages     = {4971--4980}
}

@InProceedings{Marino_2019_CVPR,
    author = {Marino, {K}enneth and {R}astegari, {M}ohammad and {F}arhadi, {A}li and {M}ottaghi, {R}oozbeh},
    title = {{OK-VQA}: A {V}isual {Q}uestion {A}nswering {B}enchmark {R}equiring {E}xternal {K}nowledge},
    booktitle = CVPR,
    year = {2019}
}

@inproceedings{zellers2019vcr,
    author = {Zellers, {R}owan and {B}isk, {Y}onatan and {F}arhadi, {A}li and {C}hoi, {Y}ejin},
    title = {From {R}ecognition to {C}ognition: {V}isual {C}ommonsense {R}easoning},
    booktitle = CVPR,
    year = {2019}
}

@InProceedings{balanced_vqa_v2,
    author = {Goyal, {Y}ash and {K}hot, {T}ejas and {S}ummers-Stay, {D}ouglas and {B}atra, {D}hruv and {P}arikh, {D}evi},
    title = {Making the {V} in {VQA} {M}atter: {E}levating the {R}ole of {I}mage {U}nderstanding in {V}isual {Q}uestion {A}nswering},
    booktitle = CVPR,
    year = {2017},
    pages     = {6325--6334},
}

@inproceedings{Bandiera2016,
    title = {A real-time system for measuring sound goodness in instrumental sounds},
    author = {Romani {P}icas, {O}riol and {P}arra {R}odriguez, {H}ector and {D}ara {D}abiri and {T}okuda, {H}iroshi and {H}ariya, {W}ataru and {O}ishi, {K}oji and {X}avier {S}erra},
    year = 2015,
    booktitle = {Proceedings of the {AES} {C}onvention}
}

@article{Fayek2019,
    title = {Temporal {R}easoning via {A}udio {Q}uestion {A}nswering},
    author = {Fayek, {H}aytham M. and {J}ohnson, {J}ustin},
    journal = IEEEAUDIO,
    pages     = {2283--2294},
    year = {2020}
}

@inproceedings{PieropanEtAl2014IROS,
    title = {Audio-{V}isual {C}lassification and {D}etection of {H}uman {M}anipulation {A}ctions},
    author = {Pieropan, {A}lessandro and {S}alvi, {G}iampiero and {P}auwels, {K}arl and {K}jellström, {H}edvig},
    year = 2014,
    booktitle = IROS,
    pages={3045--3052}
}

@inproceedings{ZhangEtAl2019AAAI,
    title = {Active {M}ini-Batch {S}ampling using {R}epulsive {P}oint {P}rocesses},
    author = {Cheng {Z}hang and {C}engiz Öztireli and {S}tephan {M}andt and {G}iampiero {S}alvi},
    year = 2019,
    booktitle = AAAI,
    pages = {5741--5748},
}

@inproceedings{CleararXiv2018,
    title = {{CLEAR}: A {D}ataset for {C}ompositional {L}anguage and {E}lementary {A}coustic {R}easoning},
    author = {Abdelnour, Jérôme and {S}alvi, {G}iampiero and {R}ouat, {J}ean},
    year = 2018,
    booktitle=NeurIPSVigil,
    note = {Available at https://arxiv.org/abs/1811.10561}
}

@inproceedings{szegedy2014going,
    title = {Going {D}eeper with {C}onvolutions},
    author = {Christian {S}zegedy and {W}ei {L}iu and {Y}angqing {J}ia and {P}ierre {S}ermanet and {S}cott {R}eed and {D}ragomir {A}nguelov and {D}umitru {E}rhan and {V}incent {V}anhoucke and {A}ndrew {R}abinovich},
    booktitle = CVPR,
    pages={1--9},
    year = {2015}
}

@inproceedings{krizhevsky2012imagenet,
    title = {Image{N}et classification with deep convolutional neural networks},
    author = {Krizhevsky, {A}lex and {S}utskever, {I}lya and {H}inton, {G}eoffrey E},
    year = 2012,
    booktitle=NeurIPS,
    pages = {1097--1105},
}

@inproceedings{simonyan2014deep,
    title = {Very {D}eep {C}onvolutional {N}etworks for {L}arge-Scale {I}mage {R}ecognition},
    author = {Karen {S}imonyan and {A}ndrew {Z}isserman},
    booktitle=ICLR,
    year = {2015}
}

@article{kumar2017deep,
    title = {Deep {CNN} {F}ramework for {A}udio {E}vent {R}ecognition using {W}eakly {L}abeled {W}eb {D}ata},
    author = {Anurag {K}umar and {B}hiksha {R}aj},
    year = 2017,
    eprint = {1707.02530},
    volume    = {abs/1707.02530},
    journal = {arXiv},
    archivePrefix = {arXiv}
}

@inproceedings{cnnAudioClass2017,
    title = {{CNN} architectures for large-scale audio classification},
    author = {Hershey, {S}hawn and {C}haudhuri, {S}ourish and {E}llis, {D}aniel P. W. and {G}emmeke, {J}ort F. and {J}ansen, {A}ren and {M}oore, R. {C}hanning and {P}lakal, {M}anoj and {P}latt, {D}evin and {S}aurous, {R}if A. and {S}eybold, {B}ryan and {S}laney, {M}alcom and {W}eiss, {R}on J. and {W}ilson, {K}evin},
    year = 2017,
  booktitle=ICASSP,
    pages = {131--135},
}

@inproceedings{Boddapati2017,
    title = {Classifying environmental sounds using image recognition networks},
    author = {Boddapati, {V}enkatesh and {P}etef, {A}ndrej and {R}asmusson, {J}im and {L}undberg, {L}ars},
    year = 2017,
    booktitle = {Procedia {C}omputer {S}cience},
    pages = {2048--2056},
    doi = {10.1016/j.procs.2017.08.250},
}

@article{ZhangEtAl2017VQAwithspeech,
    title = {Speech-Based {V}isual {Q}uestion {A}nswering},
    author = {Ted {Z}hang and {D}engxin {D}ai and {T}inne {T}uytelaars and {M}arie{-}Francine {M}oens and {L}uc {V}an {G}ool},
    year = 2017,
    journal = {arXiv},
    volume    = {abs/1705.00464},
    archivePrefix = {arXiv},
    eprint = {1705.00464}
}

@book{chang1996symbolic,
author    = {Shi{-}Kuo {C}hang and
               {E}rland {J}ungert},
  title     = {Symbolic {P}rojection for {I}mage {I}nformation {R}etrieval and {S}patial {R}easoning},
  series    = {Signal processing and its applications},
  publisher = {Elsevier},
  year      = {1996},
  url       = {https://doi.org/10.1016/b978-0-12-168030-5.x5000-1},
  doi       = {10.1016/b978-0-12-168030-5.x5000-1},
  isbn      = {978-0-12-168030-5},
}

@book{moktefi2013visual,
  title     = {Visual {R}easoning with {D}iagrams},
  series    = {Studies in {U}niversal {L}ogic},
  publisher = {Springer},
  author    = {Moktefi, Amirouche and Shin, Sun-Joo},
  year      = {2013},
  url       = {https://doi.org/10.1007/978-3-0348-0600-8},
  doi       = {10.1007/978-3-0348-0600-8},
  isbn      = {978-3-03-480599-5}
}

@article{champagne2015,
    title = {Sound reasoning (literally): {P}rospects and {C}hallenges of current acoustic logics},
    author = {Champagne, {M}arc},
    year = 2015,
    journal = {Logica {U}niversalis},
    pages = {331--343}
}

@inproceedings{champagne2018,
    title = {Teaching {A}rgument {D}iagrams to a {S}tudent {W}ho {I}s {B}lind},
    author = {Champagne, {M}arc},
    year = 2018,
    booktitle = Diagrams,
    pages = {783--786}
}

@inproceedings{antol2015vqa,
    title = {{VQA}: {V}isual {Q}uestion {A}nswering},
    author = {Antol, {S}tanislaw and {A}grawal, {A}ishwarya and {L}u, {J}iasen and {M}itchell, {M}argaret and {B}atra, {D}hruv and {L}awrence {Z}itnick, C and {P}arikh, {D}evi},
    year = 2015,
    booktitle = {ICCV},
    pages = {2425--2433},
}

@inproceedings{agrawal2016analyzing,
    title = {Analyzing the {B}ehavior of {V}isual {Q}uestion {A}nswering {M}odels},
    author = {Agrawal, {A}ishwarya and {B}atra, {D}hruv and {P}arikh, {D}evi},
    year = 2016,
    booktitle=EMNLP,
    doi = "10.18653/v1/D16-1203",
    pages = {1955--1960}
}

@inproceedings{johnson2017clevr,
    title = {{CLEVR}: A {D}iagnostic {D}ataset for {C}ompositional {L}anguage and {E}lementary {V}isual {R}easoning},
    author = {Johnson, {J}ustin and {H}ariharan, {B}harath and van der {M}aaten, {L}aurens and {F}ei-Fei, {L}i and {Z}itnick, C {L}awrence and {G}irshick, {R}oss},
    year = 2017,
    booktitle = CVPR,
    pages = {1988--1997}
}

@inproceedings{gao2015you,
    title = {Are you talking to a machine? dataset and methods for multilingual image question},
    author = {Gao, {H}aoyuan and {M}ao, {J}unhua and {Z}hou, {J}ie and {H}uang, {Z}hiheng and {W}ang, {L}ei and {X}u, {W}ei},
    year = 2015,
    booktitle=NeurIPS,
    pages = {2296--2304},
}

@inproceedings{zhu2016visual7w,
    title = {Visual7{W}: {G}rounded {Q}uestion {A}nswering in {I}mages},
    author = {Zhu, {Y}uke and {G}roth, {O}liver and {B}ernstein, {M}ichael and {F}ei-Fei, {L}i},
    year = 2016,
    booktitle = CVPR,
    pages = {4995--5004},
}

@inproceedings{zhang2016yin,
    title = {Yin and yang: {B}alancing and answering binary visual questions},
    author = {Zhang, {P}eng and {G}oyal, {Y}ash and {S}ummers-Stay, {D}ouglas and {B}atra, {D}hruv and {P}arikh, {D}evi},
    year = 2016,
    booktitle = CVPR,
    pages = {5014--5022}
}

@inproceedings{cao2005automated,
    title = {Automated question answering from lecture videos: {NLP} vs. pattern matching},
    author = {Cao, {J}inwei and {R}obles-Flores, {J}ose {A}ntonio and {R}oussinov, {D}mitri and {N}unamaker, {J}ay F},
    year = 2005,
    booktitle = HICSS,
    pages = {43b}
}

@inproceedings{chua2003question,
    title = {Question answering on large news video archive},
    author = {Chua, {T}at-Seng},
    year = 2003,
    booktitle = ISPA,
    pages = {289--294},
}

@inproceedings{kim2017deepstory,
    title = {{DeepStory}: {V}ideo story {QA} by deep embedded memory networks},
    author = {Kim, {K}yung-Min and {H}eo, {M}in-Oh and {C}hoi, {S}eong-Ho and {Z}hang, {B}young-Tak},
    booktitle = IJCAI,
    pages = {2016--2022},
    year = {2017},
    doi = {10.24963/ijcai.2017/280},
    url = {https://doi.org/10.24963/ijcai.2017/280},
}

@InProceedings{perez2017film,
    title = {{FiLM}: {V}isual {R}easoning with a {G}eneral {C}onditioning {L}ayer},
    author = {Perez, {E}than and {S}trub, {F}lorian and {D}e {V}ries, {H}arm and {D}umoulin, {V}incent and {C}ourville, {A}aron},
    booktitle = AAAI,
    year = {2018},
    pages     = {3942--3951},
}

@inproceedings{voorhees1999trec,
    title = {The {TREC}-8 {Q}uestion {A}nswering {T}rack {R}eport.},
    author = {Voorhees, {Ellen M} },
    year = 1999,
    booktitle = TREC,
    pages = {77--82}
}

@inproceedings{relu,
    title = {What is the best multi-stage architecture for object recognition?},
    author = {Jarrett, {K}evin and {K}avukcuoglu, {K}oray and {R}anzato, {M}arcAurelio and {L}eCun, {Y}ann},
    year = 2009,
    booktitle = ICCV,
    pages     = {2146--2153},
}

@inproceedings{ravichandran2002learning,
    title = {Learning surface text patterns for a question answering system},
    author = {Ravichandran, {D}eepak and {H}ovy, {E}duard},
    year = 2002,
    booktitle = ACL,
    pages = {41--47},
}

@inproceedings{hovy2000question,
    title = {Question {A}nswering in {W}ebclopedia.},
    author = {Hovy, {E}duard H and {G}erber, {L}aurie and {H}ermjakob, {U}lf and {J}unk, {M}ichael and {L}in, {C}hin-Yew},
    year = 2000,
    booktitle = TREC,
    pages = {53--56},
}

@article{Han2016AcousticScene,
    title = {Acoustic scene classification using convolutional neural network and multiple-width frequency-delta data augmentation},
    author = {Han, {Y}oonchang and {L}ee, {K}yogu},
    journal={arXiv},
    year = 2016,
    volume    = {abs/1607.02383},
    archivePrefix = {arXiv},
    eprint = {1607.02383}
}

@article{russakovsky2015imagenet,
    title = {Image{N}et large scale visual recognition challenge},
    author = {Russakovsky, {O}lga and {D}eng, {J}ia and {S}u, {H}ao and {K}rause, {J}onathan and {S}atheesh, {S}anjeev and {M}a, {S}ean and {H}uang, {Z}hiheng and {K}arpathy, {A}ndrej and {K}hosla, {A}ditya and {B}ernstein, {M}ichael and others},
    year = 2015,
    journal = IJCV,
    pages = {211--252},
}

@incollection{lecun2012efficient,
    title = {Efficient backprop},
    author = {LeCun, {Y}ann A and {B}ottou, L{\'e}on and {O}rr, {G}enevieve B and M{\"u}ller, {K}laus-Robert},
    year = 2012,
    publisher = {Springer},
    booktitle = {Neural networks: {T}ricks of the trade},
    pages = {9--48},
}

@article{Nam2019MusicGenre,
    title = {Deep {L}earning for {A}udio-Based {M}usic {C}lassification and {T}agging: {T}eaching {C}omputers to {D}istinguish {R}ock from {B}ach},
    author = {Nam, {J}uhan and {C}hoi, {K}eunwoo and {L}ee, {J}ongpil and {C}hou, {S}zu-Yu and {Y}ang, {Y}i-Hsuan},
    year = 2019,
    journal = {IEEE {S}ignal {P}rocessing {M}agazine},
    pages = {41--51},
}

@article{zheng2018cnnsbased,
    title = {{CNNs}-based {A}coustic {S}cene {C}lassification using {M}ulti-Spectrogram {F}usion and {L}abel {E}xpansions},
    author = {Weiping {Z}heng and {Z}henyao {M}o and {X}iaotao {X}ing and {G}ansen {Z}hao},
    year = 2018,
    eprint = {1809.01543},
    volume    = {abs/1809.01543},
    journal = {arXiv},
    archivePrefix = {arXiv}
}

@article{lee2017raw,
    title = {Raw {W}aveform-based {A}udio {C}lassification {U}sing {S}ample-level {CNN} {A}rchitectures},
    author = {Jongpil {L}ee and {T}aejun {K}im and {J}iyoung {P}ark and {J}uhan {N}am},
    year = 2017,
    eprint = {1712.00866},
    volume    = {abs/1712.00866},
    journal = {arXiv},
    archivePrefix = {arXiv}
}

@article{abdoli2019endtoend,
title = {End-to-end environmental sound classification using a 1{D} convolutional neural network},
journal = {Expert {S}ystems with {A}pplications},
pages = {252--263},
year = {2019},
issn = {0957-4174},
doi = {https://doi.org/10.1016/j.eswa.2019.06.040},
url = {https://www.sciencedirect.com/science/article/pii/S0957417419304403},
author = {Sajjad {A}bdoli and {P}atrick {C}ardinal and {A}lessandro {Lameiras {K}oerich}},
}

@article{CQT1992,
    title = {An efficient algorithm for the calculation of a constant {Q} transform},
    author = {Brown, {J}udith and {P}uckette, {M}iller},
    year = 1992,
    journal = JASA,
    pages = 2698,
    doi = {10.1121/1.404385}
}

@inbook{stft1987,
    title = {Short-Time {F}ourier {T}ransform},
    author = {Nawab, S. {H}amid and {Q}uatieri, {T}homas F.},
    year = 1987,
    booktitle = {Advanced {T}opics in {S}ignal {P}rocessing},
    publisher = {Prentice-Hall, {I}nc.},
    pages = {289–337},
    numpages = 49
}

@inproceedings{mfcc2000,
    title = {Mel {F}requency {C}epstral {C}oefficients for {M}usic {M}odeling},
    author = {Logan, {B}eth},
    year = 2000,
    booktitle = ISMIR,
    pages = {}
}

@inproceedings{IoffeAndSzegedy2015ICMLConditionalBatchNormalization,
 author = {Ioffe, {S}ergey and {S}zegedy, {C}hristian},
 title = {Batch {N}ormalization: {A}ccelerating {D}eep {N}etwork {T}raining by {R}educing {I}nternal {C}ovariate {S}hift},
 booktitle = ICML,
 OPTseries = {ICML'15},
 year = {2015},
 OPTlocation = {Lille, {F}rance},
 pages = {448--456},
 OPTnumpages = {9},
 url = {http://dl.acm.org/citation.cfm?id=3045118.3045167},
 OPTacmid = {3045167},
}

@inproceedings{soubbotin2001patterns,
  title={Patterns of {P}otential {A}nswer {E}xpressions as {C}lues to the {R}ight {A}nswers.},
  author={Soubbotin, {M}artin M and {S}oubbotin, {S}ergei M},
  booktitle = TREC,
  year={2001},
  volume    = {500-250},
}

@inproceedings{he2016deep,
  title={Deep {R}esidual {L}earning for {I}mage {R}ecognition},
  author={He, {K}aiming and {Z}hang, {X}iangyu and {R}en, {S}haoqing and {S}un, {J}ian},
  booktitle = CVPR,
  pages={770--778},
  year={2016}
}

@inproceedings{geman2015visual,
  title={Visual turing test for computer vision systems},
  author={Geman, {D}onald and {G}eman, {S}tuart and {H}allonquist, {N}eil and {Y}ounes, {L}aurent},
  booktitle=PNAS,
  pages={3618--3623},
  year={2015}
}

@inproceedings{yang2003videoqa,
  title={{VideoQA}: {Q}uestion {A}nswering on news video},
  author={Yang, {H}ui and {C}haisorn, {L}ekha and {Z}hao, {Y}unlong and {N}eo, {S}hi-Yong and {C}hua, {T}at-Seng},
  booktitle=MM,
  pages={632--641},
  year={2003}
}

@article{wu2008robust,
  title={A robust passage retrieval algorithm for video question answering},
  author={Wu, {Y}u-Chieh and {Y}ang, {J}ie-Chi},
  journal=IEEETCSVT,
  pages={1411--1421},
  year={2008}
}

@inproceedings{tapaswi2016movieqa,
  title={Movieqa: {U}nderstanding stories in movies through question-answering},
  author={Tapaswi, {M}akarand and {Z}hu, {Y}ukun and {S}tiefelhagen, {R}ainer and {T}orralba, {A}ntonio and {U}rtasun, {R}aquel and {F}idler, {S}anja},
  booktitle = CVPR,
  pages={4631--4640},
  year={2016}
}

@inproceedings{hudson2018compositional,
title={Compositional {A}ttention {N}etworks for {M}achine {R}easoning},
author={Drew {A}rad {H}udson and {C}hristopher D. {M}anning},
booktitle=ICLR,
year={2018},
url={https://openreview.net/forum?id=S1Euwz-Rb},
}

@inproceedings{voorhees2000building,
  title={Building a question answering test collection},
  author={Voorhees, {E}llen M and {T}ice, {D}awn M},
  booktitle=SIGIR,
  pages={200--207},
  year={2000}
}

@inproceedings{iyyer2014neural,
  title={A neural network for factoid question answering over paragraphs},
  author={Iyyer, {M}ohit and {B}oyd-Graber, {J}ordan and {C}laudino, {L}eonardo and {S}ocher, {R}ichard and {D}aum{\'e} III, {H}al},
  booktitle=EMNLP,
  pages={633--644},
  year={2014}
}

@article{Maaten2008,
author = {Maaten, {L}aurens {V}an {D}er and {H}inton, {G}eoffrey},
doi = {10.1007/s10479-011-0841-3},
journal = JMLR,
pages = {2579--2605},
title = {{Visualizing {D}ata using {t-SNE}}},
year = {2008}
}

@inproceedings{Pons2016,
author = {Pons, {J}ordi and {L}idy, {T}homas and {S}erra, {X}avier},
booktitle = CBMI,
doi = {10.1109/CBMI.2016.7500246},
isbn = {978-1-4673-8695-1},
pages = {1--6},
title = {{Experimenting with musically motivated convolutional neural networks}},
year = {2016}
}

@article{stevens1937scale,
  title={A scale for the measurement of the psychological magnitude pitch},
  author={Stevens, {S}tanley {S}mith and {V}olkmann, {J}ohn and {N}ewman, {E}dwin {B}roomell},
  journal=JASA,
  volume={8},
  number={3},
  pages={185--190},
  year={1937},
  publisher={Acoustical Society of America}
}

\begin{IEEEbiography}[{\includegraphics[width=1in,height=1.25in,clip,keepaspectratio]{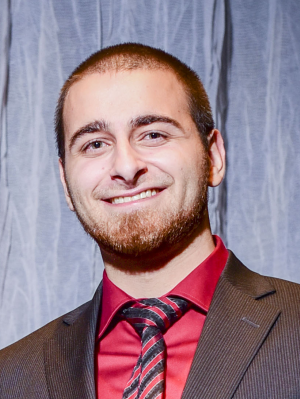}}]{Jerome Abdelnour}
is currently working on electrifying the powersport industry at Taiga Motors as a backend/devops specialist. 
He received a degree in Computer Engineering from the University of Sherbrooke (department of Electrical and Software engineering) and recently completed a Master degree in Machine Learning at the same university. His research interest include machine learning, acoustic processing, software engineering \& large-scale distributed-systems.
\end{IEEEbiography}

\begin{IEEEbiography}[{\includegraphics[width=1in,height=1.25in,clip,keepaspectratio]{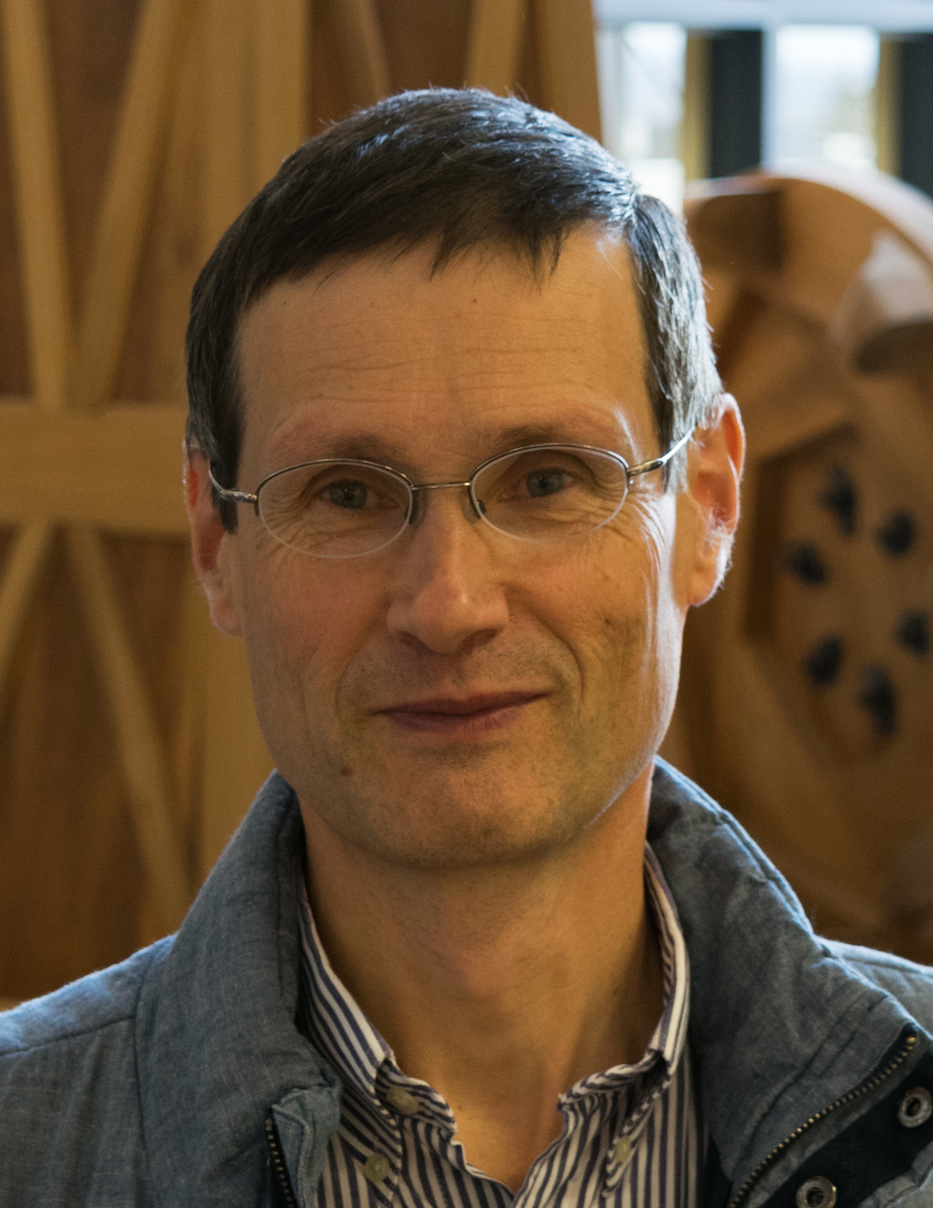}}]{Jean Rouat} Ph.D. and Full Professor is with Univ. de Sherbrooke where he founded the Computational Neuroscience and Intelligent Signal Processing Research group (NECOTIS). His translational research links neuroscience and engineering for the creation of new technologies and a better understanding of learning multimodal representations. Development of hardware low power consumption Neural Processing Units for a sustainable development, interactions with artists for multimedia and musical creations are examples of transfers that he leads based on the knowledge he gains from neuroscience. He is leading funded projects to develop sensory substitution and intelligent systems based on neuromorphic computing and implementations.
\end{IEEEbiography}

\begin{IEEEbiography}[{\includegraphics[width=1in,height=1.25in,clip,keepaspectratio]{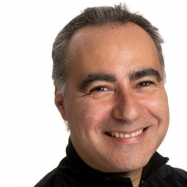}}]{Giampiero Salvi}
is Professor at the Department of Electronic Systems at the Norwegian University of Science and Technology (NTNU), Trondheim, Norway, and Associate Professor at KTH Royal Institute of Technology, Department of Electrical Engineering and Computer Science, Stockholm, Sweden. Prof. Salvi received the MSc degree in Electronic Engineering from Università la Sapienza, Rome, Italy and the PhD degree in Computer Science from KTH. 
He was a post-doctoral fellow at the Institute of Systems and Robotics, Lisbon, Portugal.
He was a co-founder of the company SynFace AB, active between 2006 and 2016.
His main interests are machine learning, speech technology, and cognitive systems.
\end{IEEEbiography}

\end{document}